\documentclass[conference]{IEEEtran}
\IEEEoverridecommandlockouts
\usepackage[bookmarks=false]{hyperref}
\usepackage{amsmath,amsfonts,amsthm} 
\usepackage{url}
\usepackage{bm}
\usepackage{xfrac}
\usepackage{stfloats}
\usepackage{fixltx2e}
\usepackage{booktabs,nicematrix}
\usepackage[caption=false,font=footnotesize]{subfig}
\usepackage{array}
\usepackage{algorithm}
\usepackage{multirow}
\usepackage{cite}
\usepackage{wrapfig}
\usepackage{lscape}
\usepackage{rotating}
\usepackage{epstopdf}
\usepackage{float}
\usepackage{algpseudocode}
\usepackage{comment}
\usepackage{stfloats}

\usepackage[normalem]{ulem}
\useunder{\uline}{\ul}{}

\usepackage{eqparbox,array}

\renewcommand{\algorithmiccomment}[1]{\bgroup\hfill\scriptsize//~#1\egroup}

\usepackage{tikz}
\usetikzlibrary{decorations.pathreplacing,calc}

\newcommand\Val{\mathcal{V}}

\begin{document}
\algnewcommand{\algorithmicgoto}{\textbf{go to}}%
\algnewcommand{\Goto}[1]{\algorithmicgoto~\ref{#1}}%
\algnewcommand\algorithmicinput{\textbf{Input:}}
\algnewcommand\Input{\item[\algorithmicinput]}
\algnewcommand\algorithmicoutput{\textbf{Output:}}
\algnewcommand\Output{\item[\algorithmicoutput]}

\title{Local overlap reduction procedure for dynamic ensemble selection}


\author{\IEEEauthorblockN{Mariana A. Souza\IEEEauthorrefmark{1},
Robert Sabourin\IEEEauthorrefmark{1},
George D. C. Cavalcanti\IEEEauthorrefmark{2} and
Rafael M. O. Cruz\IEEEauthorrefmark{1}
}
\IEEEauthorblockA{\IEEEauthorblockA{\IEEEauthorrefmark{1}\'{E}cole de Technologie Sup\'{e}rieure - Universit\'{e} du Qu\'{e}bec, Montreal, Quebec, Canada\\
Email: mariana.araujo.souza@gmail.com, robert.sabourin@etsmtl.ca, rafael.menelau-cruz@etsmtl.ca}
\IEEEauthorrefmark{2}Centro de Inform\'{a}tica - Universidade Federal de Pernambuco, Recife, Pernambuco, Brazil\\
Email: gdcc@cin.ufpe.br}}


\maketitle

\begin{abstract}
Class imbalance is a characteristic known for making learning more challenging for classification models as they may end up biased towards the majority class. A promising approach among the ensemble-based methods in the context of imbalance learning is Dynamic Selection (DS). DS techniques single out a subset of the classifiers in the ensemble to label each given unknown sample according to their estimated competence in the area surrounding the query. Because only a small region is taken into account in the selection scheme, the global class disproportion may have less impact over the system's performance. However, the presence of local class overlap may severely hinder the DS techniques' performance over imbalanced distributions as it not only exacerbates the effects of the under-representation but also introduces ambiguous and possibly unreliable samples to the competence estimation process. Thus, in this work, we propose a DS technique which attempts to minimize the effects of the local class overlap during the classifier selection procedure. The proposed method iteratively removes from the target region the instance perceived as the hardest to classify until a classifier is deemed competent to label the query sample. The known samples are characterized using instance hardness measures that quantify the local class overlap. Experimental results show that the proposed technique can significantly outperform the baseline as well as several other DS techniques, suggesting its suitability for dealing with class under-representation and overlap. Furthermore, the proposed technique still yielded competitive results when using an under-sampled, less overlapped version of the labelled sets, specially over the problems with a high proportion of minority class samples in overlap areas. Code available at \url{https://github.com/marianaasouza/lords}.
\end{abstract}

\section{Introduction}

\par Imbalanced classification problems are characterized by a disproportion in the number of instances from the problems' classes. 
Because one class is underrepresented compared to the remaining one(s), some traditional classification models may become biased towards the more well-represented labels \cite{prati2015class}. 
This bias may hinder the performance over the rarer label, which is often also the most relevant class in real-world imbalanced problems, such as fraudulent transactions detection \cite{wei2013effective} and biomedical diagnosis \cite{mazurowski2008training}. 

\par Methods specifically tailored to deal with imbalanced datasets may fall into one of the following four categories \cite{book-imbalearn}: algorithm-level approaches, which adapt traditional classifiers to deal with the class disproportion; data-level approaches, which resample the data to balance the distribution; cost-sensitive learning frameworks, which apply and incorporate class-based costs to the learning procedure; and ensemble based approaches, which usually couple ensemble methods with other methods, most commonly data-level and cost-sensitive approaches. 

\par Among the ensemble-based approaches, dynamic selection (DS) schemes were often shown to perform rather well over imbalanced distributions \cite{oliveira2017online}, specially in combination with pre-processing methods \cite{roy2018study}. 
DS techniques select a subset of the classifiers in the ensemble for labelling each query sample in particular, with the aim of using only the members that are perceived as competent in the area where the target instance is located. 
Their local approach in generalization can be an advantage in imbalanced scenarios as the global class disproportion may have 
a limited impact
over the performance, 
similarly to other local methods \cite{garcia2008k}. 

\par Also due to their local approach, however, the presence of local class overlap 
can degrade their recognition rates not only over the positive (minority) class but also the negative (majority) class \cite{prati2004class,garcia2007empirical}. 
In fact, the use of local adaptive distance and/or prototype selection methods that aim at minimizing the class overlap in the target region, called the Region of Competence (RoC) in the DS literature, were shown to improve their performance, including over highly imbalanced distributions \cite{firedes++,souza2019oneval}. 

\par Thus, in this work, we propose a dynamic selection technique which dynamically edits the target region taking into account the instances' characteristics with respect to (w.r.t.) the class overlap surrounding them. 
Based on the K-Nearest-Oracles Eliminate (KNORA-E) \cite{knora}, the proposed method also searches for local oracles, that is, classifiers in the pool that can correctly label all instances in the RoC. 
Until a competent model is found, the region is iteratively reduced by removing from it the sample with the highest estimated classification difficulty.  
Thus, the instances considered relatively more unreliable due to their ambiguity are increasingly disregarded for estimating the competence of the classifiers. 

\par The contributions of this work are then:
\begin{itemize}
    \item A novel DS technique which integrates instance characterization, including two proposed adaptations for imbalanced distributions, 
    into the definition of the RoC for dynamically reducing the class overlap in it.
    
    \item An experimental analysis over 64 imbalanced datasets in which we assess the performance of the proposed technique against 11 DS techniques, as well as the impact of the proposed dynamic overlap-reducing scheme compared to using a pre-processing technique with the same goal.
    
\end{itemize}

\par This work is organized as follows. 
Section \ref{sec:background} presents a brief background on DS techniques. 
Then, in Section \ref{sec:problem-statement} we lay out the problem statement. 
The proposed method is introduced in Section \ref{sec:proposed}. 
The experimental analysis is presented in Section \ref{sec:experimental-results}. 
Lastly, we summarize our conclusions in Section \ref{sec:conclusion}.

\section{Background}
\label{sec:background}


\par Dynamic selection techniques select a subset of a pool of base-classifiers to label each given query sample according to their perceived competence in the task, estimated over a region around the target instance. 
DS techniques are usually performed in three steps: RoC definition, competence estimation and classifier selection \cite{cruz_dynamic_2018}. 
In the first step, the RoC is defined over a labelled set, called the \textit{DSEL} set, using the nearest neighbors rule \cite{knora}, clustering methods \cite{soares2006using}, distance-based potential functions \cite{rrc}, among others. 
Then, the competence of the classifiers in the pool is estimated using the samples in the RoC according to some criteria, for instance local accuracy \cite{knora}. 
Lastly, the classifier(s) deemed competent is(are) selected to label the query. If only one the most competent one is selected, the method is a Dynamic Classifier Selection (DCS) technique while if more than one can be selected the method is a Dynamic Ensemble Selection (DES) technique, which requires a combination scheme to join the selected classifiers' responses. 

\par It has been observed in the literature that the RoC definition has a large impact over the performance of the DS techniques \cite{cruz2018prototype}, as the classifiers' competence is estimated as a function of the instances included in it. 
In fact, a few DS techniques present a RoC editing scheme with the purpose of improving the classifiers' competence estimation. 
The Multiple Classifier Behavior \cite{mcb} uses only the subset of instances from the RoC that have a similar output profile, that is, the aggregate of classifiers' responses, as the query's. 
In \cite{pereira2018dynamic}, Item Response Theory (IRT) is applied over the ensemble to filter out from a larger RoC the samples with low discrimination index. 
Lastly, the KNORA-E (and several methods based on it \cite{oliveira2018k}) also removes instances from the RoC, but in an iterative way. 
We describe the KNORA-E technique next.




\section{Problem statement}
\label{sec:problem-statement}

\par
\begin{figure}[!htb]
    \centerline{
    \includegraphics[width=0.16\textwidth]{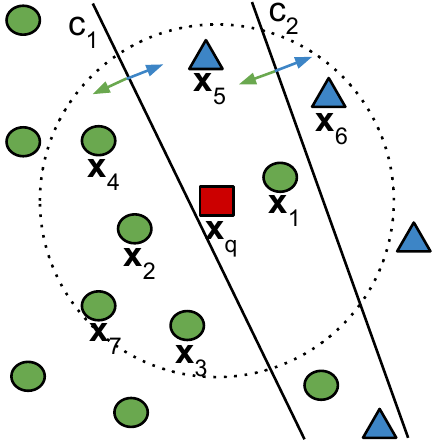}
    }
    \caption{Illustrative example. 
    $\mathbf{x}_q$ is a query instance that belongs to the locally underrepresented blue class. 
    The dashed line delimits its RoC $\theta_q = \{\mathbf{x}_1, \mathbf{x}_2, \mbox{...}
    , \mathbf{x}_7\}$, with the distances to the query $d(\mathbf{x}_q, \mathbf{x}_1) < d(\mathbf{x}_q, \mathbf{x}_2) < d(\mathbf{x}_q, \mathbf{x}_3) < d(\mathbf{x}_q, \mathbf{x}_4) < d(\mathbf{x}_q, \mathbf{x}_5) < d(\mathbf{x}_q, \mathbf{x}_6) < d(\mathbf{x}_q, \mathbf{x}_7)$. 
    The Perceptrons $c_1$ and $c_2$ form the pool of classifiers and label blue to their right and green to their left, as indicated by the arrows. 
    }
    \label{fig:original-roc}
\end{figure}

 The KNORA-E technique attempts to find in the pool a classifier that can correctly label all instances in the RoC, referred to as a local oracle. 
Fig.~\ref{fig:original-roc} shows a toy example which we use to illustrate the technique. 
It depicts the initial RoC $\theta_q = \{\mathbf{x}_1, \mathbf{x}_2, \mbox{...}, \mathbf{x}_7\}$ obtained using the K-Nearest Neighbors (KNN), the query sample $\mathbf{x}_q$ and the pool of linear classifiers $C=\{c_1, c_2\}$. 
The larger the index $i$ in $\mathbf{x}_i$, the furthest from the query the sample $\mathbf{x}_i$ is. 
The KNORA-E tries to find a classifier in $C$ that can label all $\mathbf{x}_i \in \theta_q$ correctly. 
If there is none, it removes from $\theta_q$ the instance that is the most distant from the query $\mathbf{x}_q$, and repeats the search for a local oracle. 
This process is repeated until at least one classifier is found to correctly label all remaining instances in the RoC. 
In the example from Fig.~\ref{fig:original-roc}, the RoC is edited until the removal of $\mathbf{x}_5$, as with $\theta_q = \{\mathbf{x}_1, \mathbf{x}_2, \mbox{...}, \mathbf{x}_4\}$ the classifier $c_2$ would be considered a local oracle and selected to label $\mathbf{x}_q$. 

\par As shown in this example, the RoC editing procedure from the KNORA-E may lead to the entire elimination of one of the classes in the region, which can bias the selection towards the remaining class, often the less sparse or under-represented one. 
Two methods based on the KNORA-E, the K-Nearest Oracles-Borderline (KNORA-B) and K-Nearest Oracles-Borderline-Imbalanced (KNORA-BI) propose to fix this issue by forcing the presence of all classes in the RoC. 
However, the three methods still assume the samples that are the closest to the query are more relevant for estimating the classifiers' competences. 
This may not be true if the RoC presents a certain degree of class overlap, as in the example from Fig.~\ref{fig:original-roc}. 
In fact, by disregarding the query's closest neighbor ($\mathbf{x}_1$), which is located in the most overlapped area of the RoC, the classifier $c_1$ would be recognized as a local oracle and would correctly label $\mathbf{x}_q$. 

\section{Proposed technique}
\label{sec:proposed}

\par We propose a dynamic selection technique based on the KNORA-E which performs the RoC editing based on the samples' estimated classification difficulty. 
The classification difficulty is estimated using an instance hardness measure \cite{smith2014instance,arruda2020measuring} which attempts to capture the degree of class overlap where the instance is located. 
That is, for a given unknown sample, we remove from the RoC the neighbors with higher instance hardness first, instead of the ones with largest distance to the query, in the search for the nearest local oracles. 
That way, the samples which are perceived as more reliable (that is, that are easier to classify) have a larger impact on the classifiers' competence estimation compared to the more unreliable ones in the RoC. 
Thus, the proposed neighborhood editing procedure functions similarly to a dynamic instance selection over the RoC in which the samples with higher class ambiguity are sequentially removed until a competent classifier is found. 

%

\par Algorithms \ref{alg:hardness-estimation} and \ref{alg:proposed-selection} present in more detail the proposed dynamic selection scheme. 
The instance hardness estimation step described in Algorithm \ref{alg:hardness-estimation} occurs in memorization, while the selection of the ensemble of classifiers (Algorithm \ref{alg:proposed-selection}) happens in generalization. 

\subsection{Instance hardness estimation}
\label{sec:hardness-estimation}

\par With regards to the hardness estimation procedure, Algorithm \ref{alg:hardness-estimation} requires the DSEL set and returns the hardness estimates of each sample in the DSEL. 
The hardness estimate consists of the score obtained from an instance hardness measure computed 
over the DSEL set. 
Thus, from Line 1 to Line 4, the instance hardness of each sample $\mathbf{x}_i$ in the DSEL ($\Val$) is estimated and stored. 
The instance hardness measures that are computed in $estimate\_hardness()$ are explained next. 
The hardness estimates of all instances in the DSEL are then returned in Line 5. 



\begin{algorithm}[!hb]
\centering
 \footnotesize
\begin{algorithmic}[1]
\Input $\Val = \{(\mathbf{x}_1, y_1), (\mathbf{x}_2, y_2), ..., (\mathbf{x}_N, y_N)\}$ \Comment{DSEL set}
\Output $H = \{h_1, h_2, ..., h_N\}$ \Comment{Hardness estimates}
\For {every $(\mathbf{x}_i, y_i)$ in $\Val$}
    \State $h_i \gets estimate\_hardness((\mathbf{x}_i, y_i), \Val)$ \Comment{Estimate instance hardness}
    \State $H \gets H \cup h_i$ 
\EndFor\\
\Return $H$
\end{algorithmic} 
\caption{Instance hardness estimation.}
\label{alg:hardness-estimation} 
\end{algorithm}

\par In order to characterize the hardness of the samples in the DSEL, we make use of instance hardness measures found in the literature. 
We have selected 
two measures to investigate in this work: the K-Disagreeing Neighbors (KDN) \cite{smith2014instance} and the Local Set Cardinality \cite{arruda2020measuring}. 
We chose these measures because they attempt to convey the classification difficulty associated with the local class overlap of the region where the sample is located. 
Since the dynamic selection technique is based on the concept of local oracles, the information regarding how ambiguous that region is 
may be valuable to assess the instance's reliability for the competence estimation step. 
Moreover, local class overlap was shown to correlate the most with classification difficulty on the instance level in \cite{smith2014instance}. 
Lastly, since both measures are based on sample counts, their adaptation for class imbalanced scenarios can be quite straightforward. 

\par We describe next the two chosen measures, as well as their proposed adaptation for imbalanced datasets. 
We also illustrate the hardness estimation using the example from Fig.~\ref{fig:original-roc}, and to calculate the scores of the instances within the RoC we assume the total number of samples in the DSEL set is $|\Val|=100$ and its imbalance ratio (IR), that is, the ratio between the majority and minority class sizes, is $IR=3.0$. 

\begin{figure}[!htb]
    \centerline{
    \subfloat[KDN]{
        \includegraphics[width=0.16\textwidth]{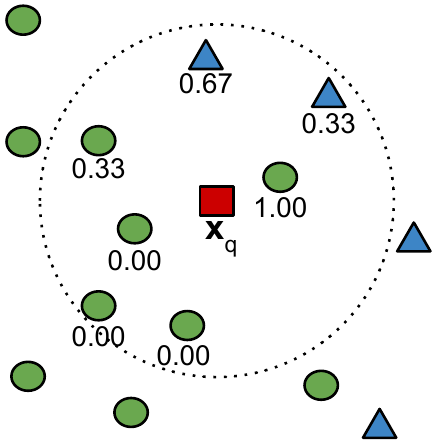}
    }
    \hspace{2.5mm}
    \hspace{2.5mm}
    \subfloat[KDNi]{
        \includegraphics[width=0.16\textwidth]{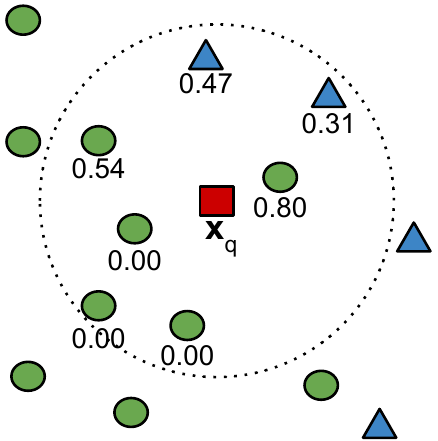}
    }
    }
    \centerline{
    \subfloat[LSC]{
        \includegraphics[width=0.16\textwidth]{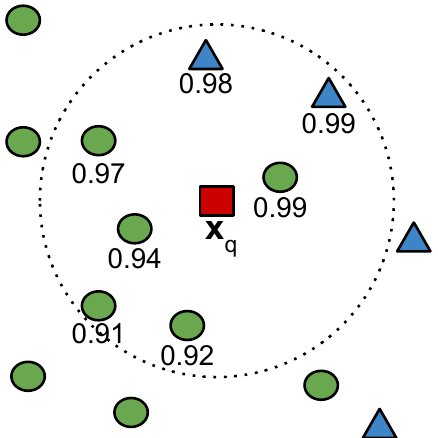}
    }
    \hspace{2.5mm}
    \hspace{2.5mm}
    \subfloat[LSCi]{
        \includegraphics[width=0.16\textwidth]{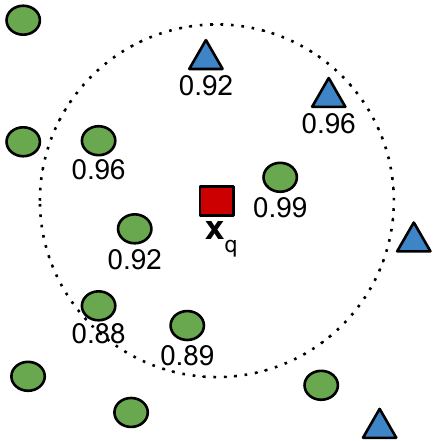}
    }
    }
    \caption{Rounded instance hardness estimates of the samples from Fig.~\ref{fig:original-roc}, according to each indicated measure.}
    \label{fig:original-roc+ih-measures}
\end{figure}

\paragraph{K-Disagreeing Neighbors (KDN)}
The KDN score indicates the proportion of neighbors of a given sample which belong to a different class. 
The computation of the KDN measure is shown in \eqref{eq:kdn}, in which the k-neighborhood $\theta_i$ of the target instance $\mathbf{x}_i$ is obtained over the remaining samples in the labelled set $\Val$. 
Fig.~\ref{fig:original-roc+ih-measures}a shows the KDN scores, with $k=3$ and rounded to two decimal points, of the instances from the toy example. 


\begin{equation}\label{eq:kdn} \footnotesize
    KDN((\mathbf{x}_i, y_i), \Val, k) = \frac{|\{(\mathbf{x}_j, y_j) \in \theta_i : y_i \neq y_j\}|}{k}  
\end{equation}

\par Since the KDN measure disregards the differences between the classes' frequencies, in highly imbalanced scenarios the scores of the positive instances in overlap regions may be estimated much higher than the negative samples. 
This could negatively impact their recognition rates, as they would be removed first in the RoC editing procedure of the proposed technique. 
Thus, we propose and evaluate an adaptation of the KDN measure which attempts to even out the scores according to the classes' frequencies. 
The adaptation, K-Disagreeing Neighbors-imbalance (KDNi) is shown in \eqref{eq:kdni}. 
It consists of the regular KDN score of the sample, according to \eqref{eq:kdn}, divided by the proportion of samples in the dataset that belong to the opposite class of the target instance ($p_{o}$). 
Since the KDN score includes the value $0.0$, we add a very small number $e=10^{-3}$ to it before computing the KDNi. 
Moreover, as the KDNi score can reach quite large values due to the division, we apply the function $f(x) = 1 - 1 / (1 + x)$ to bound the base score from \eqref{eq:kdni} and keep it in the range $(0.0, 1.0)$, with the higher values for harder instances and lower values for easier samples. 


\begin{equation}\label{eq:kdni} \footnotesize
    KDNi((\mathbf{x}_i, y_i), \Val, k) = \frac{KDN((\mathbf{x}_i, y_i), \Val, k)}{p_{o}}  
\end{equation}

\par The KDNi scores of the example from Fig.~\ref{fig:original-roc} are shown in Fig.~\ref{fig:original-roc+ih-measures}b. 
We can see, comparing the KDNi scores to the original KDN scores, that the instances very close to the border had their estimated hardness changed, while the ones further from the border changed little to nothing.

\paragraph{Local Set Cardinality (LSC)}
The LSC is an instance hardness measure based on the Local Set (LS) concept \cite{leyva2014set}. 
The LS of a given sample is comprised of the instances whose distance to it is smaller than the distance between the target sample and its nearest enemy, that is, the closest instance from the opposite class. 
The LSC measure of an instance is then the cardinality of its LS averaged by the number of samples in the dataset, as shown in \eqref{eq:lsc}, where $d()$ is the Euclidean distance and $\mathbf{x}_{ne}$ is the nearest enemy of $\mathbf{x}_i$. 
In order to yield a higher score to a harder to classify instance, we do as in \cite{lorena2018complex} and calculate the measure as one minus the score shown in \eqref{eq:lsc}. 
The LSC scores of the instances from Fig.~\ref{fig:original-roc} are shown in Fig.~\ref{fig:original-roc+ih-measures}c. 

\begin{equation}\label{eq:lsc} \footnotesize
    LSC((\mathbf{x}_i, y_i), \Val) = \frac{|\{(\mathbf{x}_j, y_j) \in \Val : d(\mathbf{x}_i, \mathbf{x}_j) < d(\mathbf{x}_i, \mathbf{x}_{ne})\}|}{|\Val|} 
\end{equation}

\par As the LSC measure also does not take into account the disproportion between the classes' sizes, we adapt it to imbalanced distributions. 
The Local Set Cardinality-imbalance is a straightforward adaptation in which instead of dividing the size of the local set by the total number of instances in the dataset $|\Val|$, we divide it by the number of samples that share the same label as the target instance, as shown in \eqref{eq:lsci}. 
As in the LSC score used in this work, we also compute the LSCi as one minus the score shown in the equation for it to have a higher value for harder samples. 

\begin{equation}\label{eq:lsci} \footnotesize
    LSCi((\mathbf{x}_i, y_i), \Val) = \frac{|\{(\mathbf{x}_j, y_j) \in \Val : d(\mathbf{x}_i, \mathbf{x}_j) < d(\mathbf{x}_i, \mathbf{x}_{ne})\}|}{|\{(\mathbf{x}_j, y_j) \in \Val : y_i = y_j)\}|} 
\end{equation}

\par Fig.~\ref{fig:original-roc+ih-measures}d shows the computed scores of the LSCi. 
We can see that the adaptation for imbalanced scenarios with this measure affected more instances than in the KDN adaptation case. 





\subsection{Ensemble selection}
\label{sec:ensemble-selection}

\par  Algorithm \ref{alg:proposed-selection} describes the dynamic ensemble of classifiers selection procedure. 
It takes as input the query sample $\mathbf{x}_q$, the region of competence (RoC) size $k$, the DSEL set, the pool of classifiers $C$ and the hardness estimates $H$ obtained in memorization, and returns the selected ensemble of classifiers $C' \subseteq C$. 
First, the RoC $\theta_q$ of size $k$ is obtained using the nearest neighbors rule in Line 1. 
Then, in Line 2, the subset of classifiers in $C$ which correctly label all instances in the RoC are singled out. 
If there are local oracles with regards to the original RoC $\theta_q$, they are returned in Line 9. 
Otherwise, the procedure enters the loop from Line 4 to Line 8 until there can be found a classifier able to correctly label all instances in the RoC. 
In Line 5, the hardness estimates of the instances in the current RoC $\theta_q'$, which starts as the original $\theta_q$, are obtained. 
The size of the current RoC $\theta_q'$ is then reduced by one in Line 6 by removing from it the instance with highest hardness estimate. 
If there are more than one instance with the maximum hardness score in the current RoC, the furthest one from the query sample is removed. 
Then, we update the subset $C'$ with the classifiers in $C$ which have $100\%$ accuracy over the reduced RoC in Line 7. 
If at least one classifier is in $C'$, then the loop is exited and the procedure returns the ensemble $C'$, which is aggregated using majority voting to produce the predicted label of the query. 
However, if no local oracle is found at that iteration, the RoC editing process is repeated with the current RoC $\theta_q'$.



\begin{algorithm}[!hb]
\centering
 \footnotesize
\begin{algorithmic}[1]
\Input $\mathbf{x}_q, k$ \Comment{Query instance, RoC size}
\Input $\Val = \{(\mathbf{x}_1, y_1), (\mathbf{x}_2, y_2), ..., (\mathbf{x}_N, y_N)\}$ \Comment{DSEL set}
\Input $C = \{c_1, c_2, ..., c_M\}$ \Comment{Pool of classifiers}
\Input $H = \{h_1, h_2, ..., h_N\}$ \Comment{Hardness estimates}
\Output $C'$ \Comment{Selected ensemble of classifiers}
\State $\theta_q \gets \{ (\mathbf{x}_i, y_i) \in KNN(\mathbf{x}_q, \Val, k)\}$ \Comment{Obtain RoC}
\State $C' \gets \{c_j \in C : \forall (\mathbf{x}_i, y_i) \in \theta_q , c_j(\mathbf{x}_i) = y_i\}$ \Comment{Select local oracles}
\State $\theta'_q \gets \theta_q$ \Comment{Current RoC}
\While {$C' = \emptyset$} 
    \State $H' \gets \{h_i \in H : (\mathbf{x}_i, y_i) \in \theta_q'\}$ \Comment{Obtain hardness of current RoC}
    \State $\theta'_q \gets \theta_q' - \{\mathbf{x}_i : h_i = max(H')\}$\Comment{Remove hardest instance}
    \State $C' \gets \{c_j \in C : \forall (\mathbf{x}_i, y_i) \in \theta_q' , c_j(\mathbf{x}_i) = y_i\}$ \Comment{Select local oracles}
\EndWhile\\
\Return $C'$
\end{algorithmic} 
\caption{Ensemble of classifiers selection.}
\label{alg:proposed-selection} 
\end{algorithm}

\par
\begin{figure}[!htb]
    \centerline{
    \subfloat[KDN, KDNi and LSCi]{
        \includegraphics[width=0.16\textwidth]{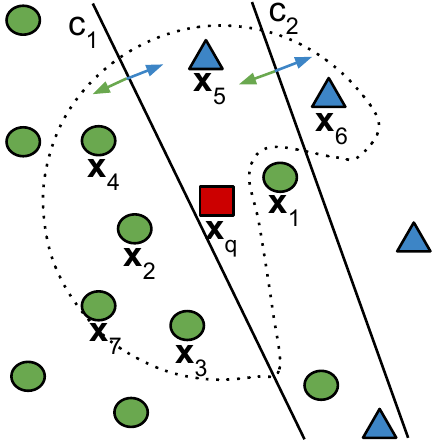}
    }
    \hspace{2.5mm}
    \hspace{2.5mm}
    \subfloat[LSC]{
        \includegraphics[width=0.16\textwidth]{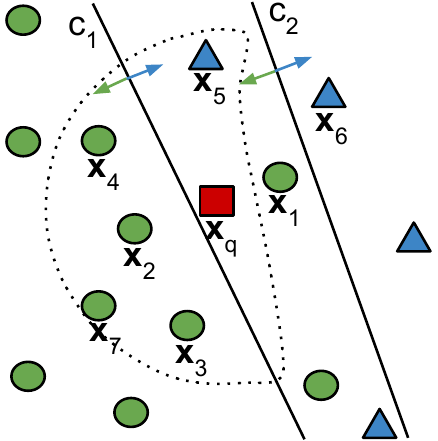}
    }
    }
    \caption{Example from Fig.~\ref{fig:original-roc} after (a) one and (b) two iteration(s) of the RoC editing loop (Line 4 to Line 8 of Algorithm \ref{alg:proposed-selection}) using the indicated measures. 
    The dotted area indicates the current RoC configuration at the time where a local oracle ($c_1$) is finally found, right after the sample $\mathbf{x}_1$ is removed.  
    }
    \label{fig:edited-roc+clfs}
\end{figure}

 Now, going back to the example from Fig.~\ref{fig:original-roc}, we can see that based on the neighboring samples' instance hardness scores obtained in memorization (Fig.~\ref{fig:original-roc+ih-measures}), the proposed technique would remove the sample $\mathbf{x}_1$ from the RoC (Line 6) in the first iteration of the loop from Line 4 to Line 8, if using the KDN, KDNi or LSCi measures. 
In this case, the edited RoC $\theta'$ obtained after the first RoC edit is shown in Fig.~\ref{fig:edited-roc+clfs}a. 
If using the LSC measure, the first sample to be removed from the original RoC would be the $\mathbf{x}_6$, as it presents the same (highest) hardness estimate as $\mathbf{x}_1$ but it is further away from the query. 
In the second iteration, however, $\mathbf{x}_1$ would be removed as as well based on the LSC scores of the remaining instances in the RoC. 
The RoC configuration obtained after the removal of $\mathbf{x}_1$ in the case of using the LSC measure in shown in Fig.~\ref{fig:edited-roc+clfs}b. 
Using any of the presented instance hardness measures, we can see from Fig.~\ref{fig:edited-roc+clfs} that, after the removal of $\mathbf{x}_1$, the method reaches its stop criteria as in Line 7 the classifier $c_1$ is identified as a local oracle in $C$, since it labels all instances in the current RoC correctly. 
Thus, the ensemble $C'$ which is returned in Line 9 contains the classifier $c_1$, which is then used to label the query instance.

\section{Experiments}
\label{sec:experimental-results}

\subsection{Experimental protocol}

\paragraph{Datasets}

\par In order to evaluate the performance of the proposed technique over class imbalanced scenarios, we use the 64 two-class datasets from the Knowledge Extraction based on Evolutionary Learning (KEEL) repository \cite{keel} presented in Table \ref{table:datasets}. 
The datasets present an IR ranging from $1.82$ to $41.40$. 
Moreover, we obtained for each dataset the proportion of safe instances according to the categorization proposed in \cite{napierala2016types} which is based on the local class distribution of the positive class samples. 
A minority class sample is considered safe if at least 4 of its 5 neighbors also belong to the minority class. 
The datasets in Table \ref{table:datasets} are sorted in descending order of S (\%), that is, the percentage of safe minority class instances. 
We refer to the first 36 problems in Table \ref{table:datasets} as safe datasets, since at least half of its positive class samples are safe ($S (\%) \geq 50$), and the remaining problems as unsafe, as in \cite{garcia2019exploring}.

\par For reproducibility, each dataset was evaluated using a stratified 5-fold cross validation procedure, one fold for test and the remaining for training, using the same partitions provided in the KEEL website. 
Due to the small-sized datasets and their high imbalance ratio, we use the training set as the DSEL set for all dynamic selection techniques evaluated, as in \cite{firedes++,souza2019oneval}. 

\begin{table}[]
\tiny
\caption{Main characteristics of the datasets used in the experiments. 
Ref. is the reference number used in this work for each dataset in the table. 
I and F refer to the number of instances and features, respectively. 
IR refers to the imbalance ratio, and 
S indicates the percentage of safe minority class instances in the whole dataset.
}
\label{table:datasets}
\setlength{\tabcolsep}{2pt}
\centerline{
\begin{NiceTabular}{@{}llcccc|llcccc@{}}
\toprule
\multicolumn{1}{l}{Ref.} & Dataset                    & \multicolumn{1}{c}{I} & \multicolumn{1}{c}{F} & \multicolumn{1}{c}{IR} & \multicolumn{1}{c}{S} & \multicolumn{1}{l}{Ref.} & Dataset                          & \multicolumn{1}{c}{I} & \multicolumn{1}{c}{F} & \multicolumn{1}{c}{IR} & \multicolumn{1}{c}{S} \\ \midrule
1                             & ecoli-0\_vs\_1             & 220                        & 7                          & 1.86                   & 99.30                      & 33                            & yeast-2\_vs\_8                   & 482                        & 8                          & 23.10                  & 55.00                      \\
2                             & shuttle-c0-vs-c4           & 1829                       & 9                          & 13.87                  & 99.19                      & 34                            & yeast-2\_vs\_4                   & 514                        & 8                          & 9.08                   & 54.90                      \\
3                             & vowel0                     & 988                        & 13                         & 9.98                   & 98.89                      & 35                            & led7digit-0-2-4-5-6-7-8-9\_vs\_1 & 443                        & 7                          & 10.97                  & 51.35                      \\
4                             & iris0                      & 150                        & 4                          & 2.00                   & 98.00                      & 36                            & shuttle-c2-vs-c4                 & 129                        & 9                          & 20.50                  & 50.00                      \\
5                             & segment0                   & 2308                       & 19                         & 6.02                   & 96.35                      & 37                            & glass1                           & 214                        & 9                          & 1.82                   & 48.68                      \\
6                             & wisconsin                  & 683                        & 9                          & 1.86                   & 91.21                      & 38                            & ecoli-0-2-6-7\_vs\_3-5           & 224                        & 7                          & 9.18                   & 45.45                      \\
7                             & vehicle2                   & 846                        & 18                         & 2.88                   & 89.45                      & 39                            & glass-0-1-6\_vs\_5               & 184                        & 9                          & 19.44                  & 44.44                      \\
8                             & page-blocks-1-3\_vs\_4     & 472                        & 10                         & 15.86                  & 82.14                      & 40                            & glass-0-4\_vs\_5                 & 92                         & 9                          & 9.22                   & 44.44                      \\
9                             & ecoli2                     & 336                        & 7                          & 5.46                   & 76.92                      & 41                            & ecoli-0-6-7\_vs\_3-5             & 222                        & 7                          & 9.09                   & 40.91                      \\
10                            & vehicle0                   & 846                        & 18                         & 3.25                   & 75.38                      & 42                            & ecoli-0-6-7\_vs\_5               & 220                        & 6                          & 10.00                  & 40.00                      \\
11                            & ecoli-0-3-4-6\_vs\_5       & 205                        & 7                          & 9.25                   & 75.00                      & 43                            & yeast6                           & 1484                       & 8                          & 41.40                  & 37.14                      \\
12                            & ecoli-0-4-6\_vs\_5         & 203                        & 6                          & 9.15                   & 75.00                      & 44                            & yeast-0-2-5-6\_vs\_3-7-8-9       & 1004                       & 8                          & 9.14                   & 34.34                      \\
13                            & ecoli-0-3-4\_vs\_5         & 200                        & 7                          & 9.00                   & 75.00                      & 45                            & yeast5                           & 1484                       & 8                          & 32.73                  & 34.09                      \\
14                            & newthyroid2                & 215                        & 5                          & 5.14                   & 74.29                      & 46                            & ecoli3                           & 336                        & 7                          & 8.60                   & 31.43                      \\
15                            & glass6                     & 214                        & 9                          & 6.38                   & 72.41                      & 47                            & glass4                           & 214                        & 9                          & 15.46                  & 30.77                      \\
16                            & ecoli-0-1-4-7\_vs\_5-6     & 332                        & 6                          & 12.28                  & 72.00                      & 48                            & pima                             & 768                        & 8                          & 1.87                   & 28.73                      \\
17                            & ecoli-0-1-3-7\_vs\_2-6     & 281                        & 7                          & 39.14                  & 71.43                      & 49                            & vehicle1                         & 846                        & 18                         & 2.90                   & 23.04                      \\
18                            & page-blocks0               & 5472                       & 10                         & 8.79                   & 70.30                      & 50                            & glass5                           & 214                        & 9                          & 22.78                  & 22.22                      \\
19                            & ecoli-0-2-3-4\_vs\_5       & 202                        & 7                          & 9.10                   & 70.00                      & 51                            & yeast1                           & 1484                       & 8                          & 2.46                   & 22.14                      \\
20                            & ecoli4                     & 336                        & 7                          & 15.80                  & 70.00                      & 52                            & vehicle3                         & 846                        & 18                         & 2.99                   & 17.45                      \\
21                            & ecoli-0-1-4-6\_vs\_5       & 280                        & 6                          & 13.00                  & 70.00                      & 53                            & yeast-0-3-5-9\_vs\_7-8           & 506                        & 8                          & 9.12                   & 16.00                      \\
22                            & ecoli-0-1\_vs\_5           & 240                        & 6                          & 11.00                  & 70.00                      & 54                            & cleveland-0\_vs\_4               & 173                        & 13                         & 12.31                  & 15.38                      \\
23                            & new-thyroid1               & 215                        & 5                          & 5.14                   & 68.57                      & 55                            & yeast-0-5-6-7-9\_vs\_4           & 528                        & 8                          & 9.35                   & 7.84                       \\
24                            & yeast-0-2-5-7-9\_vs\_3-6-8 & 1004                       & 8                          & 9.14                   & 67.68                      & 56                            & yeast4                           & 1484                       & 8                          & 28.10                  & 7.84                       \\
25                            & glass-0-1-2-3\_vs\_4-5-6   & 214                        & 9                          & 3.20                   & 66.67                      & 57                            & yeast-1\_vs\_7                   & 459                        & 7                          & 14.30                  & 6.67                       \\
26                            & ecoli-0-1-4-7\_vs\_2-3-5-6 & 336                        & 7                          & 10.59                  & 65.52                      & 58                            & haberman                         & 306                        & 3                          & 2.78                   & 6.17                       \\
27                            & ecoli-0-3-4-7\_vs\_5-6     & 257                        & 7                          & 9.28                   & 64.00                      & 59                            & yeast-1-2-8-9\_vs\_7             & 947                        & 8                          & 30.57                  & 3.33                       \\
28                            & glass0                     & 214                        & 9                          & 2.06                   & 58.57                      & 60                            & glass-0-1-5\_vs\_2               & 172                        & 9                          & 9.12                   & 0.00                       \\
29                            & ecoli-0-1\_vs\_2-3-5       & 244                        & 7                          & 9.17                   & 58.33                      & 61                            & glass-0-1-6\_vs\_2               & 192                        & 9                          & 10.29                  & 0.00                       \\
30                            & ecoli1                     & 336                        & 7                          & 3.36                   & 57.14                      & 62                            & glass2                           & 214                        & 9                          & 11.59                  & 0.00                       \\
31                            & yeast3                     & 1484                       & 8                          & 8.10                   & 55.83                      & 63                            & glass-0-1-4-6\_vs\_2             & 205                        & 9                          & 11.06                  & 0.00                       \\
32                            & glass-0-6\_vs\_5           & 108                        & 9                          & 11.00                  & 55.56                      & 64                            & yeast-1-4-5-8\_vs\_7             & 693                        & 8                          & 22.10                  & 0.00                       \\ \bottomrule
\end{NiceTabular}}
\end{table}

\paragraph{Performance measures}

\par We  evaluate the models in this work in terms of two frequently used measures for imbalance learning: the $F_1$ score \cite{fm} and the Geometric Mean (G-mean) \cite{gm}. 
For the statistical comparisons between the techniques' performances over multiple datasets, we use the pairwise Wilcoxon signed-rank test, as recommended in \cite{demvsar2006statistical}. 

\paragraph{Dynamic Selection techniques}

\par We compare our proposed technique against 11 DES techniques, including the baseline KNORA-E. 
The chosen techniques are shown in Table \ref{table:des-techniques}. 
We use their implementation from the Python library DESLib \cite{deslib}, and evaluate them with the RoC size $k=7$, and their remaining hyperparameters as the default. 
W.r.t. the proposed method, we use the same RoC size, and evaluate it using each of the hardness measures presented in Section \ref{sec:hardness-estimation}. 
The only extra hyperparameter necessary is the neighborhood size of the KDN (as in \eqref{eq:kdn} and \eqref{eq:kdni}), which we set to $k=5$ as recommended in \cite{smith2014instance}. 

\par All techniques except for the OLP use the same pool of classifiers containing 100 Perceptrons (learning rate $\alpha = 0.001$  and number of iterations $n_{iter} = 100$) that was generated using Bagging \cite{bagging}, as in \cite{souza2019oneval}. 
To obtain the output class probabilities of the base-classifiers, we apply the sigmoid function over the hyperplanes' decision function. 

\par Lastly, since the proposed technique performs a dynamic RoC definition that takes into account the degree of class overlap associated with the location of each instance, the end result may be similar to performing an instance selection/noise removal over the DSEL set. 
Thus, we also include in the analysis the use of 
the Edited Nearest Neighbors (ENN) \cite{enn} pre-processing method, available in the Python library imbalanced-learn \cite{imblearn}, over the DSEL set only. 
The ENN, in this case, removes from the DSEL set the negative class samples which contain a majority of its $k=3$ neighbors from the positive class.


\begin{table}[]
\scriptsize
\setlength{\tabcolsep}{2pt}
\caption{Dynamic Ensemble Selection (DES) techniques included in the comparative analysis.}
\label{table:des-techniques}
\centerline{
\begin{tabular}{@{}lc@{}}
\toprule
Dynamic Ensemble Selection technique                    & Reference \\ \midrule
K-Nearest Oracles Eliminate (KNORA-E)                   &    \cite{knora}       \\
K-Nearest Oracles Union (KNORA-U)                       &     \cite{knora}       \\
Dynamic Ensemble Selection Performance (DESP)           &         \cite{woloszynski2011probabilistic}  \\
Dynamic Ensemble Selection Clustering (DESC)            &          \cite{soares2006using} \\
K-Nearest Output Profiles (KNOP)                        &     \cite{cavalin2012logid}      \\
Dynamic Ensemble Selection KNN (DES-KNN)                &    \cite{soares2006using}       \\
Meta-learning for Dynamic Ensemble Selection (META-DES) &  \cite{METADES}         \\
Randomized Reference Classifier (DES-RRC)               &   \cite{rrc}        \\
Online Local Pool (OLP) & \cite{souza2019online} \\
K-Nearest Oracles-Borderline (KNORA-B)                  &    \cite{oliveira2018k}       \\ 
K-Nearest Oracles-Borderline-Imbalanced (KNORA-BI)      &   \cite{oliveira2018k}        \\ \bottomrule
\end{tabular}}
\end{table}

\subsection{Experimental results}


\par
\begin{figure}[!htb]
    \centerline{
    \subfloat[Safe]{
    \includegraphics[width=0.26\textwidth]{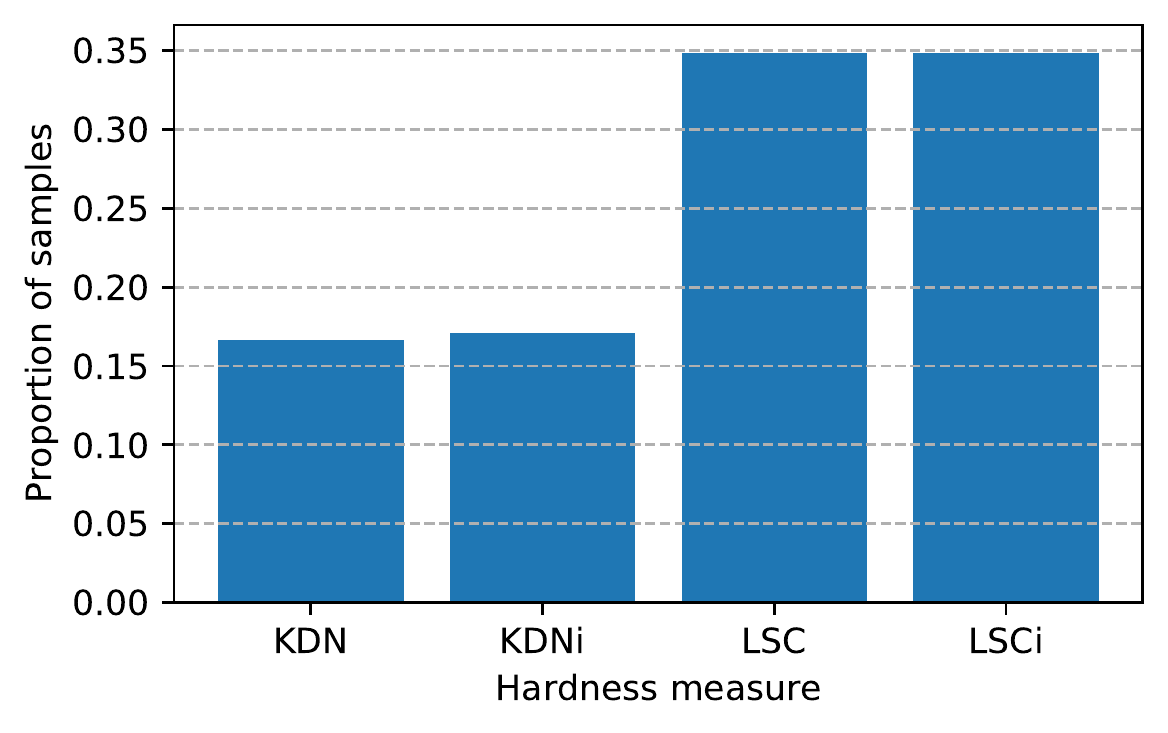}
    }
    \subfloat[Unsafe]{
    \includegraphics[width=0.26\textwidth]{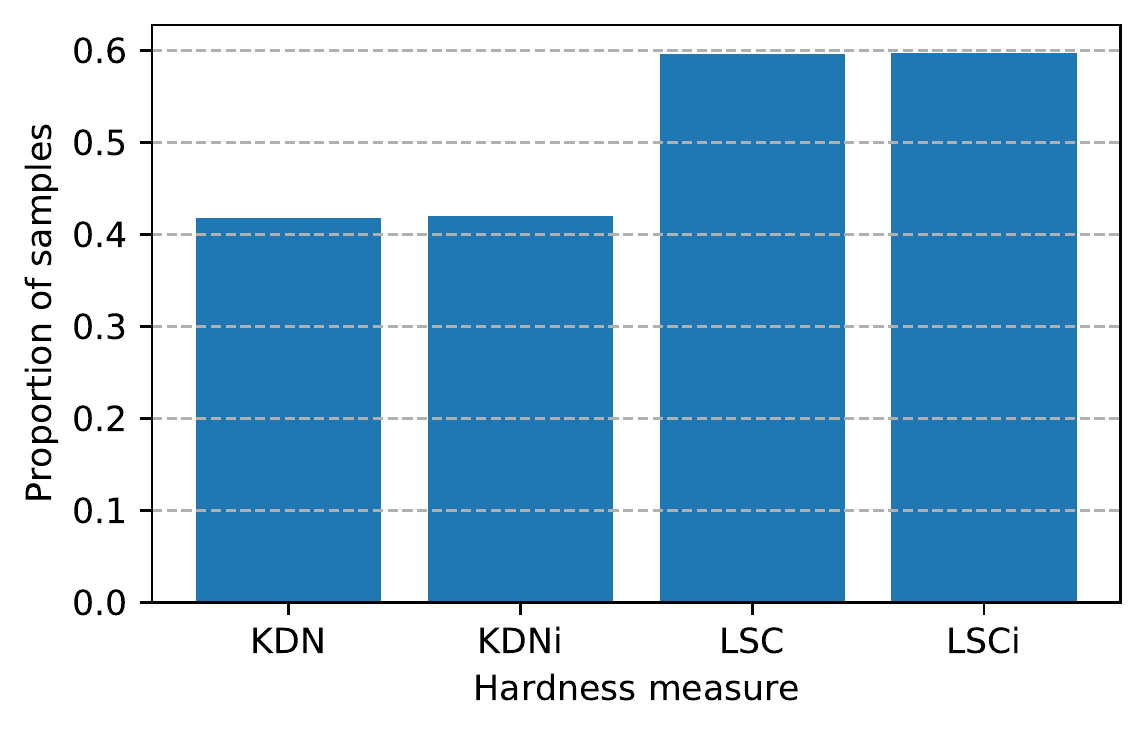}
    }
    }
    \caption{Average proportion of instances with a different sample removal order from the baseline's (KNORA-E) over the safe and unsafe datasets.}
    \label{fig:perc-samples}
\end{figure}

 We start by observing the difference in the behavior of the baseline KNORA-E and the proposed technique. 
Fig.~\ref{fig:perc-samples} shows the average proportion of test instances whose sample removal order in the RoC editing scheme was different from the distance-based order from the KNORA-E, over the safe and unsafe datasets. 
Of course, this does not necessarily mean the selected ensemble of classifiers was different than the one KNORA-E would select. 
As expected due to the higher local class overlap, changes in the sample removal order were much more frequent over the unsafe datasets compared to the safe ones, considering all measures. 
Another trend we can observe in Fig.~\ref{fig:perc-samples} is that the LSC-based measures have a larger effect on the ranking of the RoC samples compared to the KDN-based. 
This is likely due to the local set size, which can vary on a greater scale among close-by instances compared to the amount of disagreeing neighbors of a small fixed k-neighborhood. 
More score variation in a small area leads to falling fewer times on the tie rule, which is the distance-based ranking from the KNORA-E. 




\begin{figure}[!htb]
    \centerline{
    \subfloat[$F_1$]{
    \includegraphics[width=0.26\textwidth]{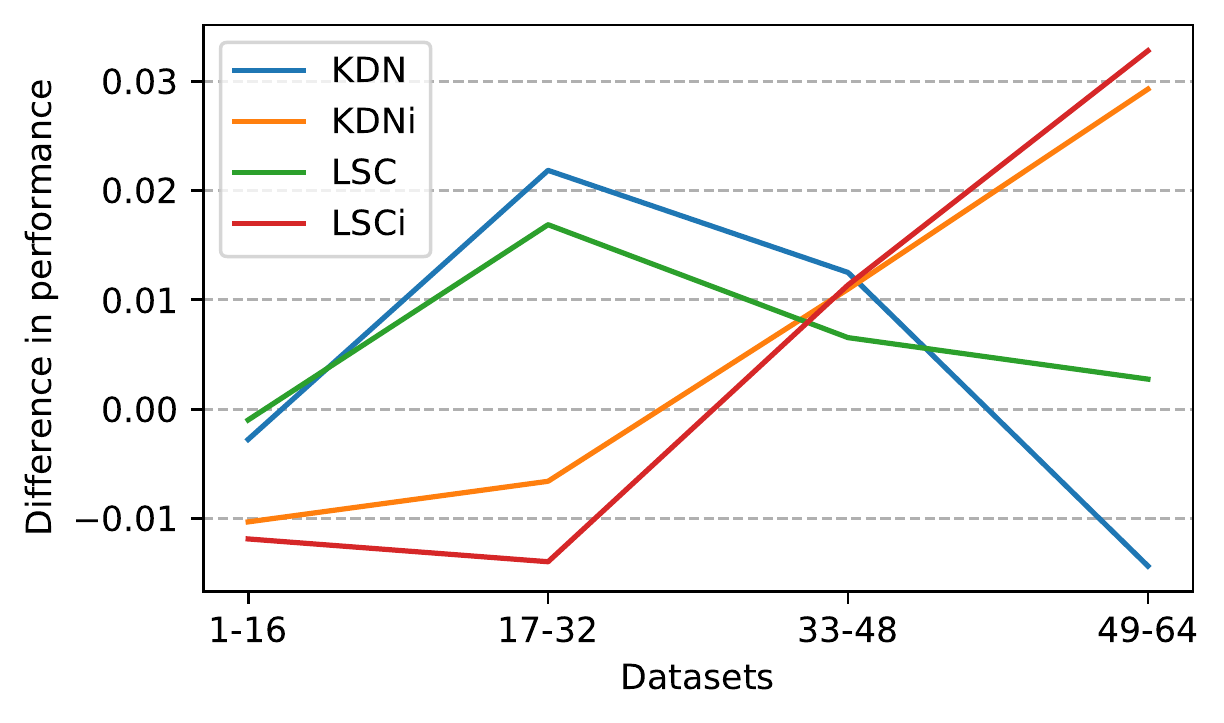}
    }
    \subfloat[G-mean]{
    \includegraphics[width=0.26\textwidth]{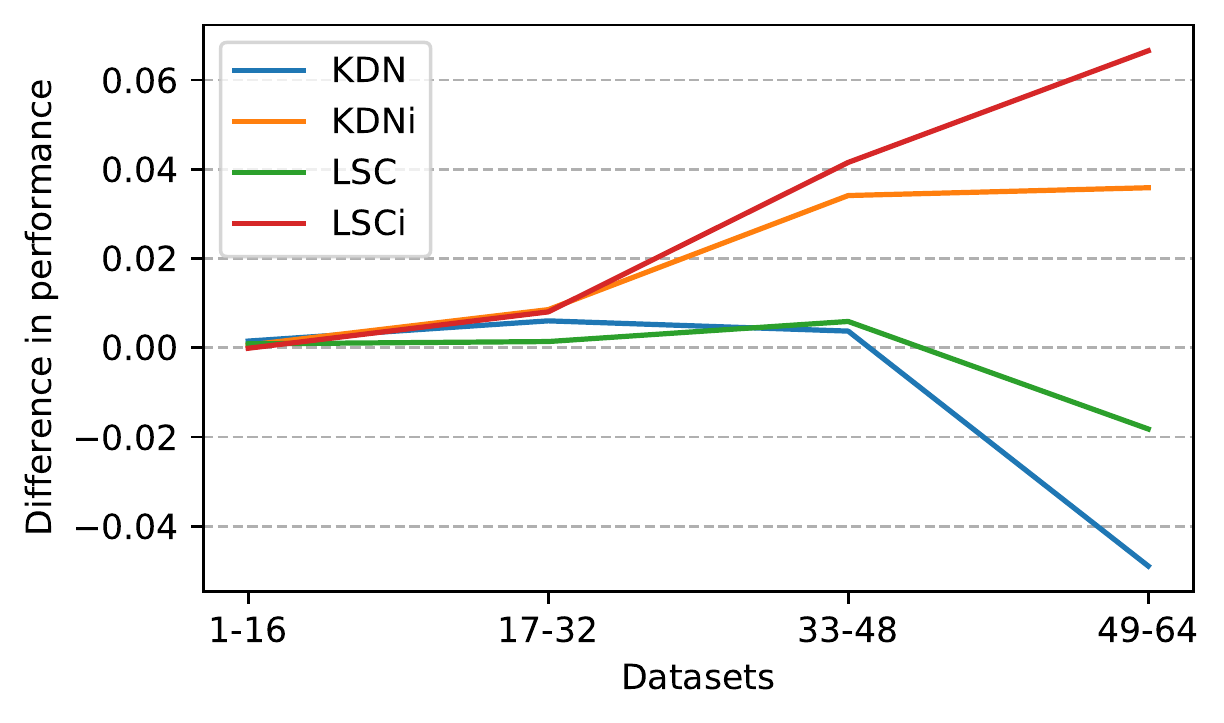}
    }
    }
    \caption{Difference in performance between the proposed method, using the indicated hardness measure, and the baseline technique (KNORA-E), averaged over the indicated datasets (Table \ref{table:datasets}). 
    }
    \label{fig:vs-kne}
\end{figure}

\par As to whether the proposed RoC editing scheme was able to outperform the KNORA-E's distance-based procedure, Fig.~\ref{fig:vs-kne} shows the difference in average performance between the proposed method, using the indicated instance hardness measure, to the baseline (KNORA-E), averaged over each range of datasets. 
It can be observed that, in terms of $F_1$, the proposed method generally outperformed the baseline using the original hardness measures (KDN and LSC)
, while using the adapted measures (KDNi and LSCi) yielded an overall superior performance over the KNORA-E over the unsafest problems
. 
This is understandable as the original measures are expected to favor the majority class instances if the minority class is more spread out. 
The adapted measures, on the other hand, favor the minority class regardless, which could explain the lower $F_1$ over the safest datasets. W.r.t. the G-mean, however, the overall performance of the proposed method using the original measures is similar or slightly better 
compared to the baseline until the last, hardest group
, at which point its performance drops relative to the KNORA-E, probably due to their bias towards the majority class in such problems.   
The gain in performance when using the adapted measures 
is visible for most groups. 




\begin{table}[!htb]
\tiny
\setlength{\tabcolsep}{2pt}
\caption{Average performance and mean rank of the DES techniques over the groups of safe and unsafe datasets. 
\textit{Wins} shows the total number of first positions. 
Solo wins count as $1$ while ties in the first position count as $1/\#$ tied techniques. 
Best results are in bold.}
\label{table:perf-ds}
\centerline{
\subfloat[Safe]{
\begin{NiceTabular}{@{}lcccc|c@{}}
\toprule
Measure      & \multicolumn{2}{c}{F1}          & \multicolumn{2}{c}{G-mean}      & \multirow{2}{*}{Wins} \\ \cmidrule(r){1-5}
Technique    & Mean           & Rank           & Mean           & Rank           &                       \\ \midrule
Prop. (KDN)  & 0.835          & 6.903          & 0.891          & 7.083          & 4.115                 \\
Prop. (KDNi) & 0.823          & 9.194          & 0.897          & 7.458          & 2.726                 \\
Prop. (LSC)  & 0.834          & 7.097          & 0.889          & 7.250          & 2.448                 \\
Prop. (LSCi) & 0.821          & 9.708          & \textbf{0.899} & 7.694          & 2.726                 \\
KNORA-E      & 0.824          & 7.917          & 0.886          & 7.958          & 1.226                 \\
KNORA-U      & 0.831          & 7.778          & 0.884          & 7.931          & 1.572                 \\
DESP         & 0.833          & 7.403          & 0.884          & 7.889          & 3.572                 \\
DESC         & 0.828          & 8.472          & 0.881          & 9.069          & 7.418                 \\
KNOP         & 0.833          & 8.306          & 0.887          & 8.681          & 1.572                 \\
DES-KNN      & 0.834          & 6.278          & 0.884          & \textbf{6.542} & 8.190                 \\
META-DES     & 0.844          & 7.861          & 0.897          & 7.542          & 3.972                 \\
DES-RRC      & 0.836          & 8.361          & 0.889          & 8.514          & 7.172                 \\
OLP          & \textbf{0.852} & \textbf{6.194} & 0.886          & 9.444          & \textbf{19.200}       \\
KNORA-B      & 0.825          & 9.111          & 0.892          & 8.819          & 2.044                 \\
KNORA-BI     & 0.818          & 9.417          & 0.895          & 8.125          & 4.044                 \\ \bottomrule
\end{NiceTabular}
}
\subfloat[Unsafe]{
\begin{NiceTabular}{@{}lcccc|c@{}}
\toprule
Measure      & \multicolumn{2}{c}{F1}          & \multicolumn{2}{c}{G-mean}      & \multirow{2}{*}{Wins} \\ \cmidrule(r){1-5}
Technique    & Mean           & Rank           & Mean           & Rank           &                       \\ \midrule
Prop. (KDN)  & 0.503          & 7.268          & 0.609          & 7.857          & 0.365                 \\
Prop. (KDNi) & \textbf{0.523} & \textbf{5.732} & 0.667          & \textbf{4.000} & 9.865                 \\
Prop. (LSC)  & 0.509          & 7.393          & 0.626          & 7.839          & 1.365                 \\
Prop. (LSCi) & \textbf{0.523} & 6.250          & \textbf{0.685} & 4.161          & \textbf{11.865}       \\
KNORA-E      & 0.508          & 7.250          & 0.636          & 6.964          & 4.143                 \\
KNORA-U      & 0.454          & 8.929          & 0.534          & 9.839          & 1.098                 \\
DESP         & 0.455          & 9.250          & 0.536          & 10.196         & 2.765                 \\
DESC         & 0.426          & 11.232         & 0.507          & 11.446         & 4.143                 \\
KNOP         & 0.444          & 9.571          & 0.533          & 9.768          & 0.765                 \\
DES-KNN      & 0.484          & 7.500          & 0.575          & 8.518          & 2.143                 \\
META-DES     & 0.507          & 6.482          & 0.613          & 6.893          & 7.098                 \\
DES-RRC      & 0.456          & 9.179          & 0.540          & 9.857          & 3.098                 \\
OLP          & 0.460          & 7.804          & 0.533          & 8.946          & 7.000                 \\
KNORA-B      & 0.500          & 8.018          & 0.635          & 7.589          & 0.143                 \\
KNORA-BI     & 0.496          & 8.143          & 0.653          & 6.125          & 0.143                 \\ \bottomrule
\end{NiceTabular}
}
}
\end{table}

\par We now include all other DES techniques from Table \ref{table:des-techniques} to the comparative analysis. 
Table \ref{table:perf-ds} shows the average $F_1$ and G-mean of each technique, as well as their average rank and total number of wins, over the safe and unsafe datasets. 
It can be observed that, over the safe datasets, the OLP yielded the highest overall $F_1$ score and average rank, with the proposed technique using the KDN and LSC obtaining the third and fourth highest scores, respectively, right after the DES-KNN. 
In terms of G-mean, the DES-KNN obtained the highest rank, followed again by the proposed method using the KDN, LSC and KDNi. 
Considering both performance measures, the OLP obtained the highest number of wins over the safe datasets, though the proposed method still obtained more wins than the baseline. 
Over the unsafe datasets, however, the proposed method obtained the two highest mean performance and average ranks in both $F_1$ and G-mean when using the adapted hardness measures, as well as the two highest number of wins. 

\begin{table}[]
\tiny
\caption{P-value obtained from the Wilcoxon signed-rank test between the average performances of the row-wise and column-wise techniques over the groups of safe and unsafe datasets. 
Values below the significance $\alpha=0.05$ are in bold. 
The symbols (+) and (-) indicate whether the column-wise (proposed) method was statistically superior or inferior, respectively, to the row-wise technique.}
\label{table:perf-ds-wilc}
\setlength{\tabcolsep}{1pt}
\centerline{
\subfloat[Safe]{
\begin{NiceTabular}{@{}lrrrr|rrrr@{}}
\toprule
Measure   & \multicolumn{4}{c|}{$F_1$}                                                                                                                           & \multicolumn{4}{c}{G-mean}                                                                                                                          \\ \midrule
Technique & \multicolumn{1}{c}{Prop. (KDN)} & \multicolumn{1}{c}{Prop. (KDNi)} & \multicolumn{1}{c}{Prop. (LSC)} & \multicolumn{1}{c|}{Prop. (LSCi)} & \multicolumn{1}{c}{Prop. (KDN)} & \multicolumn{1}{c}{Prop. (KDNi)} & \multicolumn{1}{c}{Prop. (LSC)} & \multicolumn{1}{c}{Prop. (LSCi)} \\ \midrule
KNORA-E   & 0.066                              & 0.265                               & 0.099                              & \textbf{0.025} (-)                      & \textbf{0.028} (+)                     & 0.172                               & \textbf{0.047} (+)                     & 0.364                               \\
KNORA-U   & 0.875                              & 0.296                               & 0.850                              & 0.235                               & 0.671                              & 0.282                               & 0.765                              & 0.326                               \\
DESP      & 0.648                              & 0.056                               & 0.850                              & \textbf{0.039} (-)                      & 0.354                              & 0.342                               & 0.718                              & 0.409                               \\
DESC      & 0.289                              & 0.753                               & 0.275                              & 0.561                               & 0.056                              & \textbf{0.043} (+)                      & 0.074                              & \textbf{0.041} (+)                      \\
KNOP      & 0.220                              & 0.303                               & 0.299                              & 0.174                               & 0.112                              & 0.129                               & 0.214                              & 0.129                               \\
DES-KNN   & 0.284                              & \textbf{0.004} (-)                      & 0.357                              & \textbf{0.004} (-)                      & 0.747                              & 0.632                               & 0.524                              & 0.455                               \\
META-DES  & 0.987                              & \textbf{0.007} (-)                      & 0.838                              & \textbf{0.005} (-)                      & 0.789                              & 0.747                               & 0.717                              & 0.712                               \\
DES-RRC   & 0.582                              & 0.155                               & 0.937                              & 0.094                               & 0.469                              & 0.455                               & 0.826                              & 0.474                               \\
OLP       & 0.311                              & \textbf{0.002} (-)                      & 0.146                              & \textbf{0.003} (-)                      & 0.164                              & 0.056                               & 0.248                              & \textbf{0.031}                      \\
KNORA-B   & \textbf{0.027} (+)                     & 0.862                               & \textbf{0.005} (+)                     & 0.249                               & \textbf{0.035} (+)                     & 0.170                               & \textbf{0.025} (+)                     & 0.352                               \\
KNORA-BI  & \textbf{0.028} (+)                     & 0.352                               & \textbf{0.006} (+)                     & 0.925                               & 0.250                              & 0.143                               & 0.239                              & 0.180                               \\ \bottomrule
\end{NiceTabular}}
}
\centerline{
\subfloat[Unsafe]{
\begin{NiceTabular}{@{}lrrrr|rrrr@{}}
\toprule
Measure   & \multicolumn{4}{c|}{$F_1$}                                                                                                                           & \multicolumn{4}{c}{G-mean}                                                                                                                       \\ \midrule
Technique & \multicolumn{1}{c}{Prop. (KDN)} & \multicolumn{1}{c}{Prop. (KDNi)} & \multicolumn{1}{c}{Prop. (LSC)} & \multicolumn{1}{c|}{Prop. (LSCi)} & \multicolumn{1}{c}{Prop. (KDN)} & \multicolumn{1}{c}{Prop. (KDNi)} & \multicolumn{1}{c}{Prop. (LSC)} & \multicolumn{1}{c}{Prop. (LSCi)} \\ \midrule
KNORA-E   & 0.882                              & 0.151                               & 0.546                              & 0.198                               & 0.094                              & \textbf{0.008} (+)                  & 0.657                              & \textbf{0.001} (+)                      \\
KNORA-U   & \textbf{0.034} (+)                     & \textbf{0.010} (+)                      & \textbf{0.040} (+)                     & \textbf{0.021} (+)                      & \textbf{0.004} (+)                     & \scalebox{1}{ $<$}\textbf{0.001} (+)              & \textbf{0.005} (+)                     & \scalebox{1}{ $<$}\textbf{0.001} (+)                 \\
DESP      & \textbf{0.019} (+)                     & \textbf{0.016} (+)                      & \textbf{0.043} (+)                     & \textbf{0.038} (+)                      & \textbf{0.001} (+)                     & \scalebox{1}{ $<$}\textbf{0.001} (+)              & \textbf{0.002} (+)                     & \scalebox{1}{ $<$}\textbf{0.001} (+)                 \\
DESC      & \textbf{0.001} (+)                     & \textbf{0.001} (+)                      & \textbf{0.002} (+)                     & \textbf{0.001} (+)                      & \textbf{0.001} (+)                     & \scalebox{1}{ $<$}\textbf{0.001} (+)              & \textbf{0.001} (+)                     & \scalebox{1}{ $<$}\textbf{0.001} (+)                 \\
KNOP      & \textbf{0.012} (+)                     & \textbf{0.001} (+)                      & \textbf{0.023} (+)                     & \textbf{0.002} (+)                      & \textbf{0.007} (+)                     & \scalebox{1}{ $<$}\textbf{0.001} (+)               & \textbf{0.003} (+)                     & \scalebox{1}{ $<$}\textbf{0.001} (+)                 \\
DES-KNN   & 0.172                              & 0.116                               & 0.452                              & 0.264                               & \textbf{0.029} (+)                     & \scalebox{1}{ $<$}\textbf{0.001} (+)              & \textbf{0.019} (+)                     & \scalebox{1}{ $<$}\textbf{0.001} (+)                 \\
META-DES  & 0.927                              & 0.322                               & 0.838                              & 0.682                               & 0.733                              & \textbf{0.001} (+)                   & 0.767                              & \textbf{0.001} (+)                      \\
DES-RRC   & \textbf{0.016} (+)                     & \textbf{0.010} (+)                      & \textbf{0.031} (+)                     & \textbf{0.022} (+)                      & \textbf{0.003} (+)                     & \scalebox{1}{ $<$}\textbf{0.001} (+)              & \textbf{0.004} (+)                     & \tiny{$<$}\textbf{0.001} (+)                 \\
OLP       & 0.202                              & \textbf{0.024} (+)                      & 0.096                              & \textbf{0.048} (+)                      & \textbf{0.024} (+)                     & \scalebox{1}{ $<$}\textbf{0.001} (+)              & \textbf{0.007} (+)                     & \scalebox{1}{ $<$}\textbf{0.001} (+)                 \\
KNORA-B   & 0.802                              & 0.062                               & 0.412                              & \textbf{0.038} (+)                      & 0.227                              & \textbf{0.004} (+)                   & 0.909                              & \textbf{0.001} (+)                      \\
KNORA-BI  & 0.793                              & \textbf{0.026} (+)                      & 0.339                              & \textbf{0.012} (+)                      & \textbf{0.005} (+)                     & \textbf{0.006} (+)                   & \textbf{0.006} (+)                     & \textbf{0.001} (+)                      \\ \bottomrule
\end{NiceTabular}
}
}
\end{table}

\par Table \ref{table:perf-ds-wilc} shows the p-values obtained from the pairwise Wilcoxon signed-rank test with significance $\alpha=0.05$, performed over the average performances of the DS techniques. 
First, we can see that compared to the baseline, the proposed technique yielded a statistically similar $F_1$ score over the safe datasets except when using the LSCi measure, which obtained a significantly worse performance. 
As observed previously, the LSCi prompts a change in the RoC editing order much more often than the KDN-based measures, all the while favoring the positive class, so over the safe problems the precision may have been hindered. 
In fact, both adapted measures obtained a poorer $F_1$ score compared to the DES-KNN, META-DES and OLP over the safe datasets. 
However, the proposed method using the original measures was significantly better over the safe datasets than the KNORA-B and KNORA-BI. 
Moreover, in terms of G-mean, the proposed technique significantly outperformed both the baseline and KNORA-B over the safe datasets, using the KDN and LSC as hardness estimates.

\par Over the unsafe datasets, the proposed technique using the adapted measures obtained a statistically similar $F_1$ and a significantly better G-mean compared to the baseline, META-DES and DES-KNN.  
It also significantly surpassed the OLP, KNORA-B and KNORA-BI when using the LSCi measure in both $F_1$ and G-mean. 
Compared to the remaining techniques, the proposed method significantly outperformed them using all of the hardness measures investigated. 
This suggests that reducing the local overlap in the RoC in generalization may be advantageous for the classifier selection step, specially over the unsafe datasets. 

\begin{figure}[!htb]
    \centerline{
    \subfloat[$F_1$]{
    \includegraphics[width=0.26\textwidth]{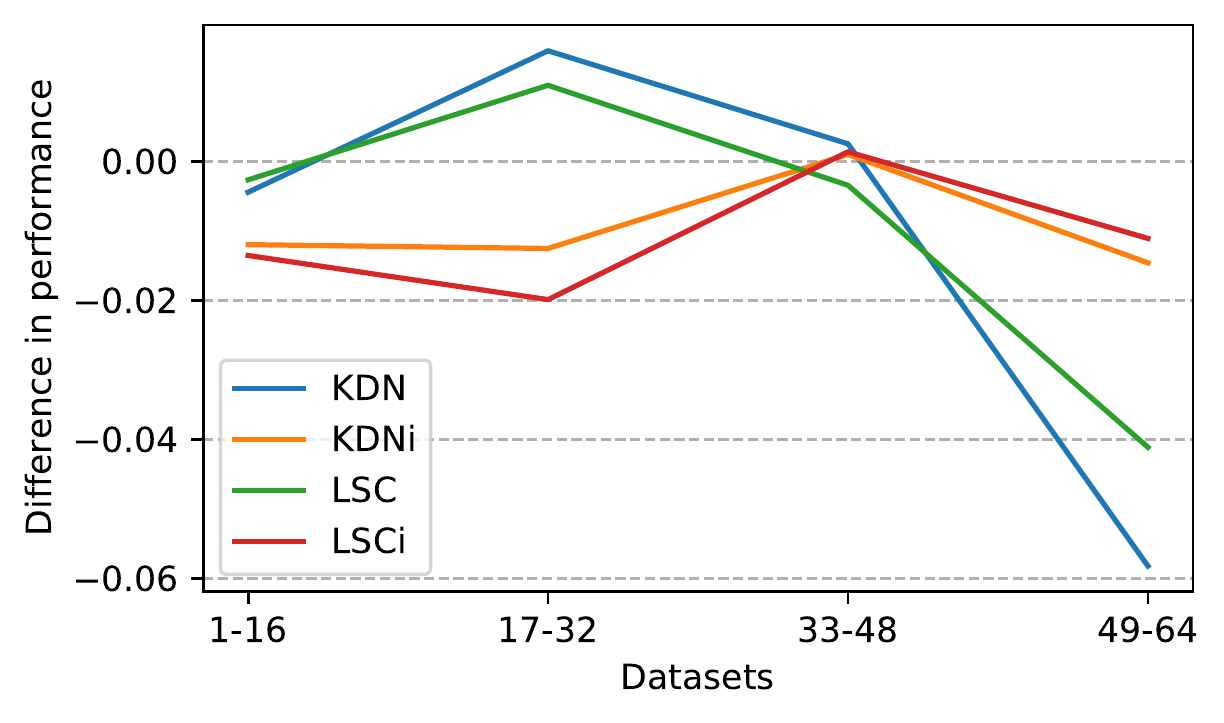}
    }
    \subfloat[G-mean]{
    \includegraphics[width=0.26\textwidth]{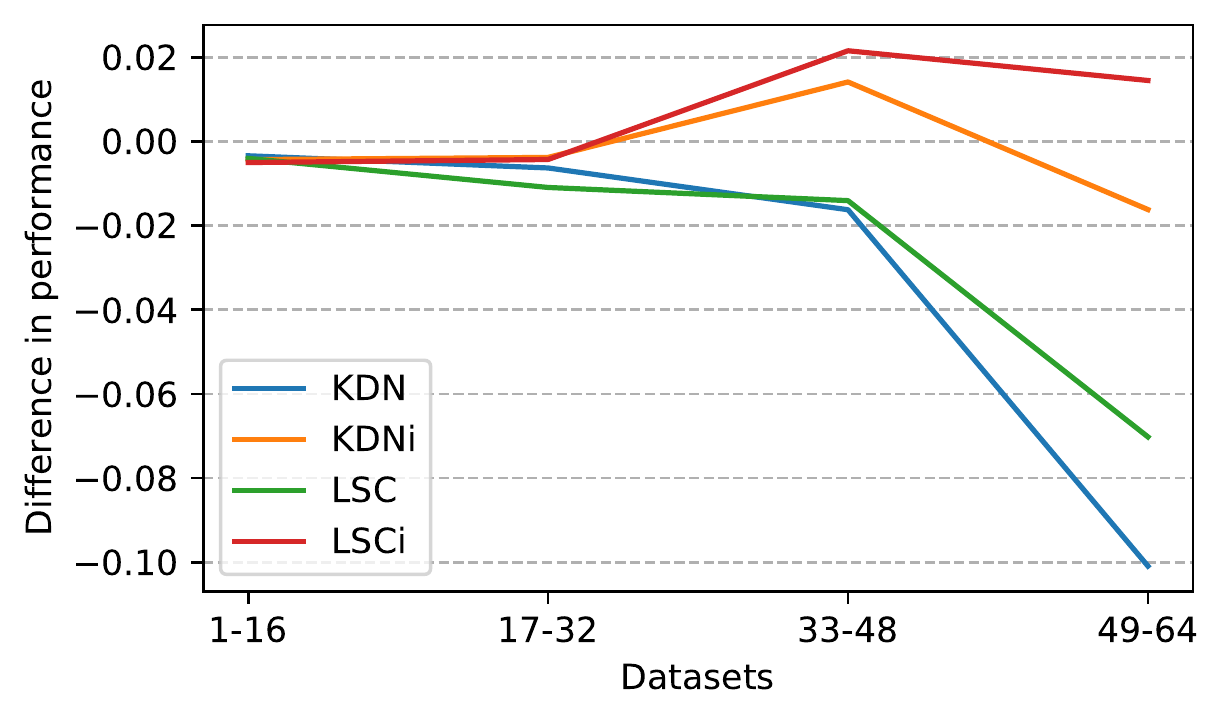}
    }
    }
    \centerline{
    \subfloat[$F_1$]{
    \includegraphics[width=0.26\textwidth]{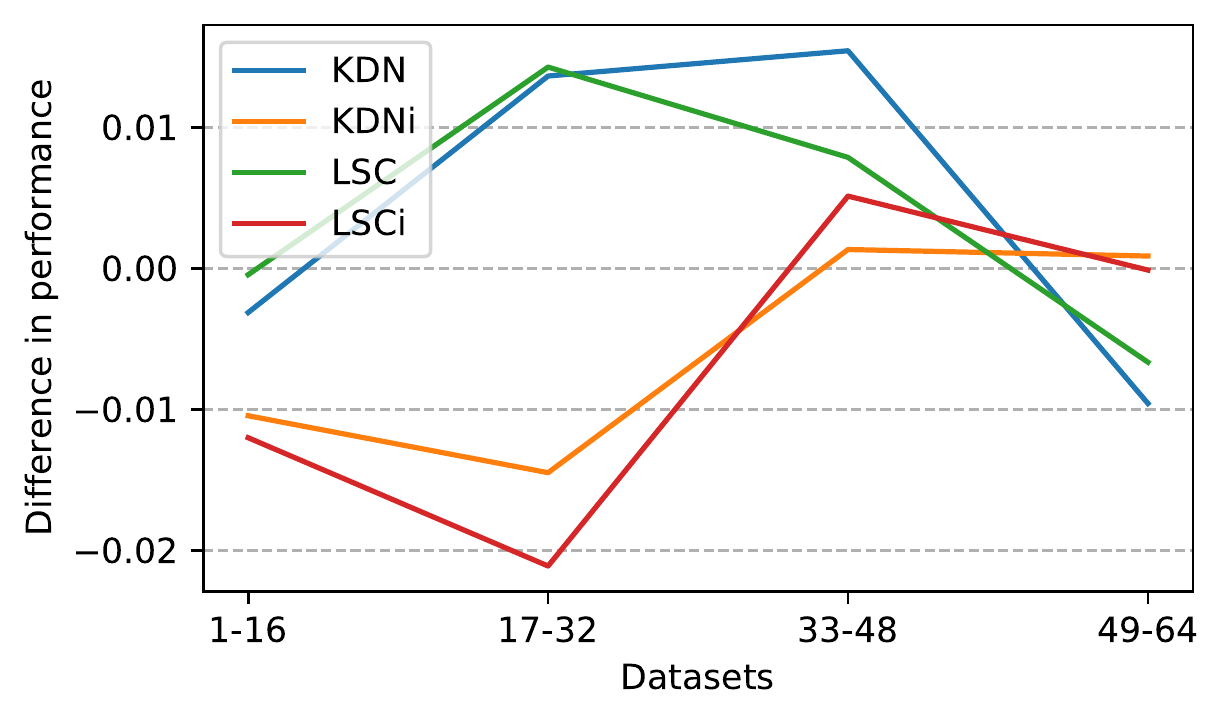}
    }
    \subfloat[G-mean]{
    \includegraphics[width=0.26\textwidth]{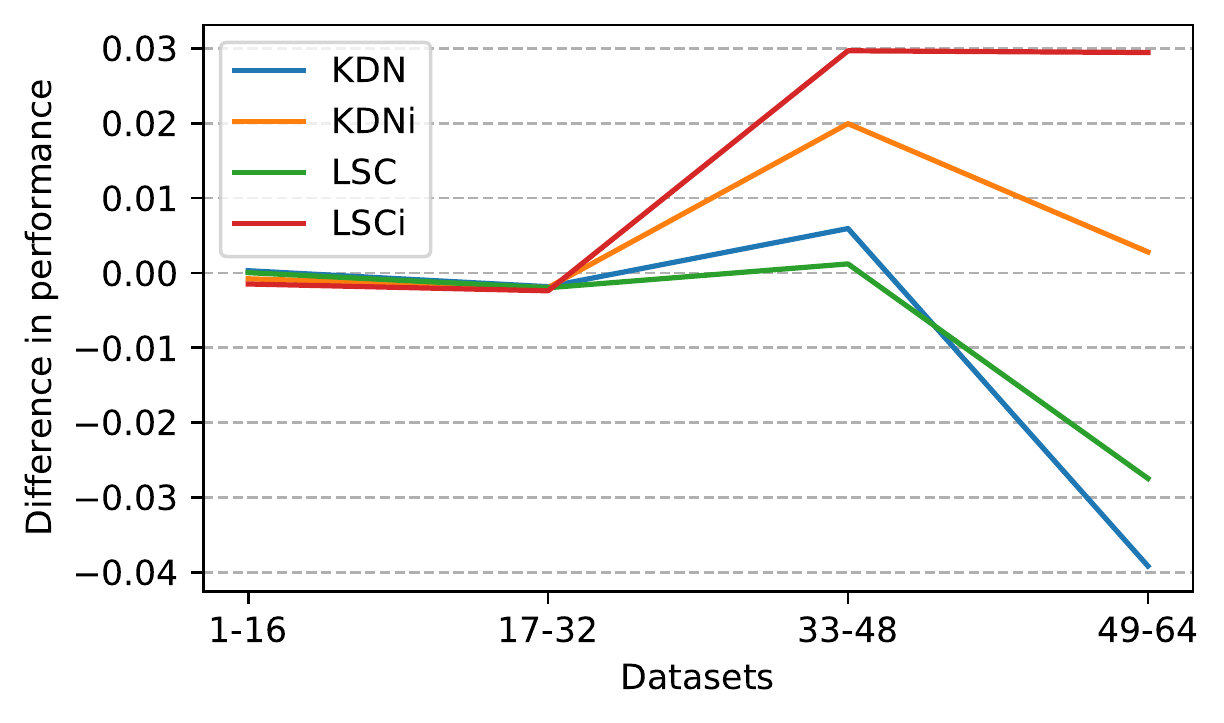}
    }
    }
     \caption{Difference in performance between the proposed method using the indicated hardness measure, (a-b) without and (b-c) with pre-processing (ENN), and the baseline technique with pre-processing (KNORA-E + ENN), averaged over the indicated datasets (Table \ref{table:datasets}). 
    }
    \label{fig:vs-kne-enn-only-all}
\end{figure}

\par We now analyze the performance of the proposed technique relative to the use of the ENN. 
In order to observe the impact of such procedure over the DS techniques, we only apply it to the DSEL set, thus the pool of classifiers is the same as in the previous analysis. 
Fig.~\ref{fig:vs-kne-enn-only-all}a and Fig.~\ref{fig:vs-kne-enn-only-all}b show the difference in average $F_1$ and G-mean between the proposed method \textit{without} pre-processing and the baseline \textit{with} pre-processing (KNORA-E + ENN). 
We can observe that applying the ENN over the DSEL yielded a greater improvement to the KNORA-E method compared to the proposed hardness-based RoC editing procedure. 
This may be due to the fact that, after applying the ENN over the DSEL, the initial RoC contains more reliable samples since the most unreliable ones in that area were already removed. 
Therefore, even if the proposed technique eventually removes such samples, comparatively fewer instances will remain in the RoC which may negatively impact the competence estimation. 
Thus, starting off the ensemble selection procedure with a less overlapped RoC seems to provide a better overall performance for the KNORA-E. 
However, for some datasets the dynamic sample removal used in the proposed method still yielded slightly better average results, which may motivate the design of an approach that can combine this desirable characteristic from a pre-processing procedure for an online RoC editing scheme.


\par Fig.~\ref{fig:vs-kne-enn-only-all}c and Fig.~\ref{fig:vs-kne-enn-only-all}d show the difference in the mean performances of the proposed method and the baseline, both using the ENN over the DSEL set now. 
We can see that the difference in performance presents similar characteristics to the results obtained for both techniques without the pre-processing (Fig.~\ref{fig:vs-kne}), but with less improvement in 
general. 
With less local overlap in the DSEL, it can be expected that the proposed RoC editing scheme will have a reduced impact in performance. 





\begin{table}[!htb]
\tiny
\caption{Average performance and mean rank of the DES techniques with the ENN pre-processing method over the groups of safe and unsafe datasets. 
\textit{Wins} shows the total number of first positions. 
Solo wins count as $1$ while ties in the first position count as $1/\#$ tied techniques. 
Best results are in bold.}
\label{table:perf-ds-enn}
\setlength{\tabcolsep}{2pt}
\centerline{
\subfloat[Safe]{
\begin{NiceTabular}{@{}lcccc|c@{}}
\toprule
Measure      & \multicolumn{2}{c}{F1}          & \multicolumn{2}{c}{G-mean}      & \multirow{2}{*}{Wins} \\ \cmidrule(r){1-5}
Technique    & Mean           & Rank           & Mean           & Rank           &                       \\ \midrule
Prop. (KDN)  & 0.838          & 6.639          & 0.897          & \textbf{6.847} & 3.218                 \\
Prop. (KDNi) & 0.823          & 9.639          & 0.900          & 7.931          & 1.468                 \\
Prop. (LSC)  & 0.837          & 6.750          & 0.895          & \textbf{6.847} & 3.318                 \\
Prop. (LSCi) & 0.822          & 10.167         & \textbf{0.903} & 8.125          & 3.068                 \\
KNORA-E      & 0.828          & 7.292          & 0.894          & \textbf{6.847} & 5.468                 \\
KNORA-U      & 0.832          & 8.306          & 0.886          & 8.875          & 2.092                 \\
DESP         & 0.832          & 8.042          & 0.886          & 8.681          & 4.092                 \\
DESC         & 0.825          & 9.347          & 0.885          & 9.542          & 2.105                 \\
KNOP         & 0.836          & 8.139          & 0.891          & 8.708          & 4.592                 \\
DES-KNN      & 0.830          & 6.611          & 0.885          & 7.236          & 5.175                 \\
META-DES     & 0.845          & 8.083          & 0.900          & 8.014          & 7.890                 \\
DES-RRC      & 0.840          & 7.681          & 0.894          & 8.139          & 7.425                 \\
OLP          & \textbf{0.852} & \textbf{5.500} & 0.890          & 8.792          & \textbf{17.486}       \\
KNORA-B      & 0.832          & 8.319          & \textbf{0.903} & 7.611          & 1.801                 \\
KNORA-BI     & 0.823          & 9.486          & 0.902          & 7.806          & 2.801                 \\ \bottomrule
\end{NiceTabular}
}
\subfloat[Unsafe]{
\begin{NiceTabular}{@{}lcccc|c@{}}
\toprule
Measure      & \multicolumn{2}{c}{F1}          & \multicolumn{2}{c}{G-mean}      & \multirow{2}{*}{Wins} \\ \cmidrule(r){1-5}
Technique    & Mean           & Rank           & Mean           & Rank           &    \multicolumn{1}{c}{}   \\ \midrule
Prop. (KDN)  & 0.535          & \textbf{6.250} & 0.653          & 6.464          & 1.254                 \\
Prop. (KDNi) & 0.531          & 6.714          & 0.680          & \textbf{4.464} & 9.254                 \\
Prop. (LSC)  & 0.535          & 6.518          & 0.659          & 6.482          & 1.254                 \\
Prop. (LSCi) & 0.530          & 7.321          & \textbf{0.697} & 4.500          & 6.254                 \\
KNORA-E      & \textbf{0.537} & 6.839          & 0.676          & 5.839          & 4.487                 \\
KNORA-U      & 0.465          & 10.107         & 0.549          & 11.518         & 0.654                 \\
DESP         & 0.478          & 8.982          & 0.564          & 10.250         & 0.654                 \\
DESC         & 0.462          & 10.339         & 0.558          & 10.768         & 5.000                 \\
KNOP         & 0.458          & 10.054         & 0.551          & 10.875         & 0.654                 \\
DES-KNN      & 0.520          & 7.107          & 0.623          & 8.232          & 1.154                 \\
META-DES     & 0.526          & 7.304          & 0.633          & 8.089          & 8.654                 \\
DES-RRC      & 0.472          & 9.500          & 0.559          & 10.946         & 2.654                 \\
OLP          & 0.498          & 6.661          & 0.579          & 8.464          & \textbf{11.100}       \\
KNORA-B      & 0.529          & 7.893          & 0.673          & 7.000          & 1.487                 \\
KNORA-BI     & 0.518          & 8.411          & 0.684          & 6.107          & 1.487                 \\ \bottomrule
\end{NiceTabular}
}}
\end{table}

\par Applying the ENN over the DSEL for all DS techniques investigated in this work, we obtain the results from Table \ref{table:perf-ds-enn}. 
We can see that the OLP still yielded the best average rank over the safe datasets in terms of $F_1$, as well as the highest number of wins considering both performance measures, but the proposed method (with the KDN and LSC) and the baseline KNORA-E obtained the highest rank in terms of G-mean. 
Over the unsafe datasets, the proposed method obtained the two highest average ranks w.r.t. $F_1$ using the original measures, and w.r.t. G-mean using the adapted measures. 
The second highest number of wins was also achieved by the proposed technique, using the KDNi. 
Interestingly using the original measures yielded a better $F_1$ than using the adapted measures over the unsafe datasets, as opposed to the previous results without the pre-processing, possibly due to the slightly less biased hardness estimation after removing the most overlapped majority class samples.


    

\begin{table}[!htb]
\tiny
\caption{P-value obtained from the Wilcoxon signed-rank test between the average performances of the column-wise and row-wise techniques, with the ENN pre-processing method, over the groups of safe and unsafe datasets. 
Values below the significance $\alpha=0.05$ are in bold. 
The symbols (+) and (-) indicate whether the column-wise (proposed) method was statistically superior or inferior, respectively, to the row-wise technique.}
\label{table:perf-ds-enn-wilc}
\setlength{\tabcolsep}{1pt}
\centerline{
\subfloat[Safe]{
\begin{NiceTabular}{@{}lrrrr|rrrr@{}}
\toprule
Measure   & \multicolumn{4}{c|}{$F_1$}                                                                                                                               & \multicolumn{4}{c}{G-mean}                                                                                                                          \\ \midrule
Technique & \multicolumn{1}{c}{Prop. (KDN)} & \multicolumn{1}{c}{Prop. (KDNi)} & \multicolumn{1}{c}{Prop. (LSC)} & \multicolumn{1}{c|}{Prop. (LSCi)}     & \multicolumn{1}{c}{Prop. (KDN)} & \multicolumn{1}{c}{Prop. (KDNi)} & \multicolumn{1}{c}{Prop. (LSC)} & \multicolumn{1}{c}{Prop. (LSCi)} \\ \midrule
KNORA-E   & 0.404                              & \textbf{0.006} (-)                      & 0.297                              & \scalebox{1}{ $<$}\textbf{0.001} (-)                     & 0.860                              & 0.057                               & 0.974                              & 0.061                               \\
KNORA-U   & 0.366                              & 0.195                               & 0.535                              & 0.133                                   & \textbf{0.039} (+)                     & 0.072                               & 0.133                              & 0.091                               \\
DESP      & 0.582                              & 0.052                               & 0.620                              & \textbf{0.036} (-)                          & 0.107                              & 0.248                               & 0.142                              & 0.342                               \\
DESC      & \textbf{0.049} (+)                     & 0.912                               & \textbf{0.046} (+)                     & 0.683                                   & \textbf{0.011} (+)                     & \textbf{0.034} (+)                      & \textbf{0.010} (+)                     & \textbf{0.026} (+)                      \\
KNOP      & 0.370                              & 0.111                               & 0.268                              & 0.072                                   & 0.076                              & 0.133                               & 0.063                              & 0.155                               \\
DES-KNN   & 0.740                              & \textbf{0.003} (-)                      & 0.918                              & \textbf{0.003} (-)                          & 0.249                              & 0.887                               & 0.138                              & 0.813                               \\
META-DES  & 0.706                              & \textbf{0.007} (-)                      & 0.274                              & \textbf{0.003} (-)                          & 0.409                              & 0.556                               & 0.404                              & 0.545                               \\
DES-RRC   & 0.881                              & \textbf{0.049} (-)                      & 0.994                              & \textbf{0.030} (-)                          & 0.465                              & 0.666                               & 0.620                              & 0.700                               \\
OLP       & 0.104                              & \textbf{0.001} (-)                      & 0.072                              & \textbf{0.002} (-)                          & 0.164                              & 0.134                               & 0.212                              & 0.126                               \\
KNORA-B   & \textbf{0.033} (+)                     & \textbf{0.012} (-)                      & \textbf{0.022} (+)                     & \scalebox{1}{ $<$}\textbf{0.001} (-) & 0.318                              & 0.623                               & 0.358                              & 0.358                               \\
KNORA-BI  & \textbf{0.018} (+)                     & 0.893                               & \textbf{0.013} (+)                     & 0.492                                   & 0.303                              & 0.663                               & 0.281                              & 0.281                               \\ \bottomrule
\end{NiceTabular}
}
}
\centerline{
\subfloat[Unsafe]{
\begin{NiceTabular}{@{}lrrrr|rrrr@{}}
\toprule
Measure   & \multicolumn{4}{c|}{$F_1$}                                                                                                                           & \multicolumn{4}{c}{G-mean}                                                                                                                          \\ \midrule
Technique & \multicolumn{1}{c}{Prop. (KDN)} & \multicolumn{1}{c}{Prop. (KDNi)} & \multicolumn{1}{c}{Prop. (LSC)} & \multicolumn{1}{c|}{Prop. (LSCi)} & \multicolumn{1}{c}{Prop. (KDN)} & \multicolumn{1}{c}{Prop. (KDNi)} & \multicolumn{1}{c}{Prop. (LSC)} & \multicolumn{1}{c}{Prop. (LSCi)} \\ \midrule
KNORA-E   & 0.793                              & 0.501                               & 0.546                              & 0.130                               & 0.161                              & 0.269                               & 0.168                              & \textbf{0.029} (+)                      \\
KNORA-U   & \textbf{0.001} (+)                     & \textbf{0.011} (+)                      & \textbf{0.001} (+)                     & \textbf{0.026} (+)                      & \scalebox{1}{ $<$}\textbf{0.001} (+)                & \scalebox{1}{ $<$}\textbf{0.001} (+)                 & \scalebox{1}{ $<$}\textbf{0.001} (+)                & \scalebox{1}{ $<$}\textbf{0.001} (+)                 \\
DESP      & \textbf{0.005} (+)                     & \textbf{0.029} (+)                      & \textbf{0.010} (+)                     & 0.056                               & \scalebox{1}{ $<$}\textbf{0.001} (+)                & \scalebox{1}{ $<$}\textbf{0.001} (+)                 & \scalebox{1}{ $<$}\textbf{0.001} (+)                & \scalebox{1}{ $<$}\textbf{0.001} (+)                 \\
DESC      & \textbf{0.001} (+)                     & \textbf{0.003} (+)                      & \textbf{0.001} (+)                     & \textbf{0.009} (+)                      & \scalebox{1}{ $<$}\textbf{0.001} (+)                & \scalebox{1}{ $<$}\textbf{0.001} (+)                 & \scalebox{1}{ $<$}\textbf{0.001} (+)                & \scalebox{1}{ $<$}\textbf{0.001} (+)                 \\
KNOP      & \textbf{0.001} (+)                     & \textbf{0.002} (+)                      & \textbf{0.003} (+)                     & \textbf{0.008} (+)                      & \scalebox{1}{ $<$}\textbf{0.001} (+)                & \scalebox{1}{ $<$}\textbf{0.001} (+)                 & \scalebox{1}{ $<$}\textbf{0.001} (+)                & \scalebox{1}{ $<$}\textbf{0.001} (+)                 \\
DES-KNN   & 0.190                              & 0.445                               & 0.241                              & 0.707                               & \textbf{0.006} (+)                     & \scalebox{1}{ $<$}\textbf{0.001} (+)                     & \textbf{0.003} (+)                     & \scalebox{1}{ $<$}\textbf{0.001} (+)                    \\
META-DES  & 0.439                              & 0.982                               & 0.600                              & 0.820                               & 0.106                              & \textbf{0.008} (+)                      & 0.106                              & \textbf{0.002} (+)                      \\
DES-RRC   & \textbf{0.003} (+)                     & 0.053                               & \textbf{0.005} (+)                     & 0.068                               & \scalebox{1}{ $<$}\textbf{0.001} (+)                & \scalebox{1}{ $<$}\textbf{0.001} (+)                 & \scalebox{1}{ $<$}\textbf{0.001} (+)                & \scalebox{1}{ $<$}\textbf{0.001} (+)                 \\
OLP       & 0.285                              & 0.316                               & 0.412                              & 0.750                               & \textbf{0.006} (+)                     & \textbf{0.001} (+)                      & \textbf{0.005} (+)                     & \textbf{0.001} (+)                      \\
KNORA-B   & 0.305                              & 0.927                               & 0.316                              & 0.733                               & 0.716                              & 0.179                               & 0.733                              & \textbf{0.029} (+)                      \\
KNORA-BI  & 0.136                              & 0.206                               & 0.136                              & 0.259                               & 0.101                              & 0.592                               & 0.127                              & 0.094                               \\ \bottomrule
\end{NiceTabular}
}
}
\end{table}

\par Table \ref{table:perf-ds-enn-wilc} shows the resulting p-values of the Wilcoxon signed-rank test on the average performances of the techniques with the ENN over the safe and unsafe datasets. 
We can observe that, over the safe datasets, the proposed method with the adapted measures was statistically worse than several methods including the baseline while with the original measures it surpassed the KNORA-B and KNORA-BI in terms of $F_1$. 
Over the unsafe datasets, however, the proposed method with the LSCi statistically outperformed all techniques except for the KNORA-BI, in terms of G-mean.  

\subsection{Lessons learned}

\par All in all, the results suggest that there is a performance improvement to be had by characterizing the hardness of the instances in the RoC, and prioritizing the ones that seem more reliable for the classifiers' competence estimation. 
Using an overlap reducing pre-processing technique tailored for imbalanced data seems to have an overall better impact over the classifier selection procedure compared to the proposed dynamic RoC editing scheme alone, possibly due to the region being less ambiguous from the start. 
However, we observed an advantage in not outright removing the instances in memorization for some datasets. 
Moreover, the proposed technique still presented a performance improvement after reducing the local class overlap in the data using the pre-processing technique, which further supports the integration of instance characterization into the RoC definition for DS techniques.

\par Furthermore, we observed that the instance characterization within the proposed method had a large effect on performance, so choosing which hardness measure to use should be based on the distribution of the positive class in the problem. 
For a minority class composed of mostly safe samples, the  original measures seemed to provide a good estimate of the instances' reliability in the region, even in the presence of high global imbalance ratios. 
On the other hand, over the unsafe datasets, the adapted measures were a much more effective guide in the proposed RoC editing scheme. 
Lastly, while the corresponding KDN-based and LS-based measures yielded somewhat similar performances, likely because they both attempt to quantify the local class overlap using a concept of neighborhood, we would recommend using the LS-based measures as they not only presented slightly better overall results but also do not require any hyperparameter.


\section{Conclusion}
\label{sec:conclusion}

\par In this work, we proposed a Dynamic Selection technique which dynamically edits the target region in the search for a local oracle. 
Motivated by the observation that a class-overlapped region can hinder the system's recognition rates, specially over the locally under-represented class, the proposed method removes the samples perceived as most unreliable from the RoC one by one until at least one classifier can label all remaining instances in it. 
For characterizing the known instances in the problem, we use two instance hardness measures that convey the degree of local overlap in the area. 
We also propose and evaluate an adapted version of these measures as they can often be biased towards the majority class. 


\par Experiments were conducted over 64 imbalanced datasets, which were split into two groups according to the percentage of safe positive class instances in the problem. 
The proposed method yielded very competitive results against a baseline and 10 other DS techniques, specially over the unsafe datasets, which suggests that the overlap-reducing procedure on the RoC can improve the competence estimation and thus the selection of the classifiers. 
Moreover, the most adequate instance characterization to use within the proposed technique appears to depend on the positive class distribution, as the original instance hardness measures present a high bias towards the majority class on the unsafe problems. 

\par Future work in this line of research may involve a further investigation on the impact of using pre-processing methods to remove the local overlap for DS techniques, as opposed to dynamically editing the RoC. 
This may lead to the study of a scheme which can provide some advantages from both approaches, in order to improve the competence estimation of the classifiers and thus the DS techniques' recognition rates over all classes. 



\section*{Acknowledgments}

The authors would like to thank the Canadian agencies FRQ (Fonds de Recherche du Qu\'{e}bec) and NSERC (Natural Sciences and Engineering Research Council of Canada), and the Brazilian agencies CAPES (Coordena\c{c}\~{a}o de Aperfei\c{c}oamento de Pessoal de N\'{i}vel Superior), CNPq (Conselho Nacional de Desenvolvimento Cient\'{i}fico e Tecnol\'{o}gico) and FACEPE (Funda\c{c}\~{a}o de Amparo \`{a} Ci\^{e}ncia e Tecnologia de Pernambuco).

\bibliographystyle{IEEEtran}
\bibliography{references}

\begin{thebibliography}{10}
\providecommand{\url}[1]{#1}
\csname url@samestyle\endcsname
\providecommand{\newblock}{\relax}
\providecommand{\bibinfo}[2]{#2}
\providecommand{\BIBentrySTDinterwordspacing}{\spaceskip=0pt\relax}
\providecommand{\BIBentryALTinterwordstretchfactor}{4}
\providecommand{\BIBentryALTinterwordspacing}{\spaceskip=\fontdimen2\font plus
\BIBentryALTinterwordstretchfactor\fontdimen3\font minus
  \fontdimen4\font\relax}
\providecommand{\BIBforeignlanguage}[2]{{%
\expandafter\ifx\csname l@#1\endcsname\relax
\typeout{** WARNING: IEEEtran.bst: No hyphenation pattern has been}%
\typeout{** loaded for the language `#1'. Using the pattern for}%
\typeout{** the default language instead.}%
\else
\language=\csname l@#1\endcsname
\fi
#2}}
\providecommand{\BIBdecl}{\relax}
\BIBdecl

\bibitem{prati2015class}
R.~C. Prati, G.~E. Batista, and D.~F. Silva, ``Class imbalance revisited: a new
  experimental setup to assess the performance of treatment methods,''
  \emph{Knowledge and Information Systems}, vol.~45, no.~1, pp. 247--270, 2015.

\bibitem{wei2013effective}
W.~Wei, J.~Li, L.~Cao, Y.~Ou, and J.~Chen, ``Effective detection of
  sophisticated online banking fraud on extremely imbalanced data,''
  \emph{World Wide Web}, vol.~16, no.~4, pp. 449--475, 2013.

\bibitem{mazurowski2008training}
M.~A. Mazurowski, P.~A. Habas, J.~M. Zurada, J.~Y. Lo, J.~A. Baker, and G.~D.
  Tourassi, ``Training neural network classifiers for medical decision making:
  The effects of imbalanced datasets on classification performance,''
  \emph{Neural networks}, vol.~21, no. 2-3, pp. 427--436, 2008.

\bibitem{book-imbalearn}
A.~Fern\'{a}ndez, S.~Garc\'{i}a, M.~Galar, R.~C. Prati, B.~Krawczyk, and
  F.~Herrera, \emph{Learning from Imbalanced Data Sets}.\hskip 1em plus 0.5em
  minus 0.4em\relax Springer International Publishing, 2018.

\bibitem{oliveira2017online}
D.~V. Oliveira, G.~D. Cavalcanti, and R.~Sabourin, ``Online pruning of base
  classifiers for dynamic ensemble selection,'' \emph{Pattern Recognition},
  vol.~72, pp. 44--58, 2017.

\bibitem{roy2018study}
A.~Roy, R.~M. Cruz, R.~Sabourin, and G.~D. Cavalcanti, ``A study on combining
  dynamic selection and data preprocessing for imbalance learning,''
  \emph{Neurocomputing}, vol. 286, pp. 179--192, 2018.

\bibitem{garcia2008k}
V.~Garc{\'\i}a, R.~A. Mollineda, and J.~S. S{\'a}nchez, ``On the k-nn
  performance in a challenging scenario of imbalance and overlapping,''
  \emph{Pattern Analysis and Applications}, vol.~11, no. 3-4, pp. 269--280,
  2008.

\bibitem{prati2004class}
R.~C. Prati, G.~E. Batista, and M.~C. Monard, ``Class imbalances versus class
  overlapping: an analysis of a learning system behavior,'' in \emph{Mexican
  international conference on artificial intelligence}.\hskip 1em plus 0.5em
  minus 0.4em\relax Springer, 2004, pp. 312--321.

\bibitem{garcia2007empirical}
V.~Garc{\'\i}a, J.~S{\'a}nchez, and R.~Mollineda, ``An empirical study of the
  behavior of classifiers on imbalanced and overlapped data sets,'' in
  \emph{Iberoamerican Congress on Pattern Recognition}.\hskip 1em plus 0.5em
  minus 0.4em\relax Springer, 2007, pp. 397--406.

\bibitem{firedes++}
R.~M. Cruz, D.~V. Oliveira, G.~D. Cavalcanti, and R.~Sabourin, ``Fire-des++:
  Enhanced online pruning of base classifiers for dynamic ensemble selection,''
  \emph{Pattern Recognition}, vol.~85, pp. 149--160, 2019.

\bibitem{souza2019oneval}
M.~A. Souza, G.~D. Cavalcanti, R.~M. Cruz, and R.~Sabourin, ``On evaluating the
  online local pool generation method for imbalance learning,'' in
  \emph{International Joint Conference on Neural Networks (IJCNN), 2019}.\hskip
  1em plus 0.5em minus 0.4em\relax IEEE, 2019, pp. 1--8.

\bibitem{knora}
A.~H.-R. Ko, R.~Sabourin, and A.~de~Souza Britto~Jr, ``A new dynamic ensemble
  selection method for numeral recognition,'' in \emph{7th International
  Conference on Multiple Classifier Systems}.\hskip 1em plus 0.5em minus
  0.4em\relax Springer-Verlag, 2007, pp. 431--439.

\bibitem{cruz_dynamic_2018}
R.~M.~O. Cruz, R.~Sabourin, and G.~D.~C. Cavalcanti, ``Dynamic classifier
  selection: {Recent} advances and perspectives,'' \emph{Information Fusion},
  vol.~41, pp. 195--216, May 2018.

\bibitem{soares2006using}
R.~G. Soares, A.~Santana, A.~M. Canuto, and M.~C.~P. de~Souto, ``Using accuracy
  and diversity to select classifiers to build ensembles,'' in \emph{The 2006
  IEEE International Joint Conference on Neural Network Proceedings}.\hskip 1em
  plus 0.5em minus 0.4em\relax IEEE, 2006, pp. 1310--1316.

\bibitem{rrc}
T.~Woloszynski and M.~Kurzynski, ``A probabilistic model of classifier
  competence for dynamic ensemble selection,'' \emph{Pattern Recognition},
  vol.~44, no.~10, pp. 2656--2668, 2011.

\bibitem{cruz2018prototype}
R.~M. Cruz, R.~Sabourin, and G.~D. Cavalcanti, ``Prototype selection for
  dynamic classifier and ensemble selection,'' \emph{Neural Computing and
  Applications}, vol.~29, no.~2, pp. 447--457, 2018.

\bibitem{mcb}
G.~Giacinto, F.~Roli, and G.~Fumera, ``Selection of classifiers based on
  multiple classifier behaviour,'' in \emph{Joint IAPR International Workshops
  on Statistical Techniques in Pattern Recognition and Structural and Syntactic
  Pattern Recognition}.\hskip 1em plus 0.5em minus 0.4em\relax Springer-Verlag,
  2000, pp. 87--93.

\bibitem{pereira2018dynamic}
M.~Pereira, A.~Britto, L.~Oliveira, and R.~Sabourin, ``Dynamic ensemble
  selection by k-nearest local oracles with discrimination index,'' in
  \emph{2018 IEEE 30th International conference on tools with artificial
  intelligence (ICTAI)}.\hskip 1em plus 0.5em minus 0.4em\relax IEEE, 2018, pp.
  765--771.

\bibitem{oliveira2018k}
D.~V. Oliveira, G.~D. Cavalcanti, T.~N. Porpino, R.~M. Cruz, and R.~Sabourin,
  ``K-nearest oracles borderline dynamic classifier ensemble selection,'' in
  \emph{2018 International Joint Conference on Neural Networks (IJCNN)}.\hskip
  1em plus 0.5em minus 0.4em\relax IEEE, 2018, pp. 1--8.

\bibitem{smith2014instance}
M.~R. Smith, T.~Martinez, and C.~Giraud-Carrier, ``An instance level analysis
  of data complexity,'' \emph{Machine Learning}, vol.~95, no.~2, pp. 225--256,
  2014.

\bibitem{arruda2020measuring}
J.~L. Arruda, R.~B. Prud{\^e}ncio, and A.~C. Lorena, ``Measuring instance
  hardness using data complexity measures,'' in \emph{Brazilian Conference on
  Intelligent Systems}.\hskip 1em plus 0.5em minus 0.4em\relax Springer, 2020,
  pp. 483--497.

\bibitem{leyva2014set}
E.~Leyva, A.~Gonz{\'a}lez, and R.~Perez, ``A set of complexity measures
  designed for applying meta-learning to instance selection,'' \emph{IEEE
  Transactions on Knowledge and Data Engineering}, vol.~27, no.~2, pp.
  354--367, 2014.

\bibitem{lorena2018complex}
A.~C. Lorena, L.~P. Garcia, J.~Lehmann, M.~C. Souto, and T.~K. Ho, ``How
  complex is your classification problem? a survey on measuring classification
  complexity,'' \emph{arXiv preprint arXiv:1808.03591}, 2018.

\bibitem{keel}
J.~Alcal{\'a}, A.~Fern{\'a}ndez, J.~Luengo, J.~Derrac, S.~Garc{\'\i}a,
  L.~S{\'a}nchez, and F.~Herrera, ``{KEEL data-mining software tool: data set
  repository, integration of algorithms and experimental analysis framework},''
  \emph{Journal of Multiple-Valued Logic and Soft Computing}, vol.~17, no. 2-3,
  pp. 255--287, 2011.

\bibitem{napierala2016types}
K.~Napierala and J.~Stefanowski, ``Types of minority class examples and their
  influence on learning classifiers from imbalanced data,'' \emph{Journal of
  Intelligent Information Systems}, vol.~46, no.~3, pp. 563--597, 2016.

\bibitem{garcia2019exploring}
V.~Garc{\'\i}a, A.~I. Marqu{\'e}s, and J.~S. S{\'a}nchez, ``Exploring the
  synergetic effects of sample types on the performance of ensembles for credit
  risk and corporate bankruptcy prediction,'' \emph{Information Fusion},
  vol.~47, pp. 88--101, 2019.

\bibitem{fm}
C.~van Rijsbergen, \emph{Information Retrieval}.\hskip 1em plus 0.5em minus
  0.4em\relax Butterworth, 1979.

\bibitem{gm}
M.~Kubat, S.~Matwin \emph{et~al.}, ``Addressing the curse of imbalanced
  training sets: one-sided selection,'' in \emph{Icml}, vol.~97.\hskip 1em plus
  0.5em minus 0.4em\relax Nashville, USA, 1997, pp. 179--186.

\bibitem{demvsar2006statistical}
J.~Dem{\v{s}}ar, ``Statistical comparisons of classifiers over multiple data
  sets,'' \emph{Journal of Machine learning research}, vol.~7, no. Jan, pp.
  1--30, 2006.

\bibitem{deslib}
R.~M. Cruz, L.~G. Hafemann, R.~Sabourin, and G.~D. Cavalcanti, ``Deslib: A
  dynamic ensemble selection library in python,'' \emph{arXiv preprint
  arXiv:1802.04967}, 2018.

\bibitem{bagging}
L.~Breiman, ``{Bagging predictors},'' \emph{Machine Learning}, vol.~24, no.~2,
  pp. 123--140, 1996.

\bibitem{enn}
D.~L. Wilson, ``Asymptotic properties of nearest neighbor rules using edited
  data,'' \emph{IEEE Transactions on Systems, Man, and Cybernetics}, no.~3, pp.
  408--421, 1972.

\bibitem{imblearn}
G.~Lema{{\^i}}tre, F.~Nogueira, and C.~K. Aridas, ``Imbalanced-learn: A python
  toolbox to tackle the curse of imbalanced datasets in machine learning,''
  \emph{Journal of Machine Learning Research}, vol.~18, no.~17, pp. 1--5, 2017.

\bibitem{woloszynski2011probabilistic}
T.~Woloszynski and M.~Kurzynski, ``A probabilistic model of classifier
  competence for dynamic ensemble selection,'' \emph{Pattern Recognition},
  vol.~44, no. 10-11, pp. 2656--2668, 2011.

\bibitem{cavalin2012logid}
P.~R. Cavalin, R.~Sabourin, and C.~Y. Suen, ``Logid: An adaptive framework
  combining local and global incremental learning for dynamic selection of
  ensembles of hmms,'' \emph{Pattern recognition}, vol.~45, no.~9, pp.
  3544--3556, 2012.

\bibitem{METADES}
R.~M.~O. Cruz, R.~Sabourin, G.~D.~C. Cavalcanti, and T.~I. Ren, ``{META-DES: A
  dynamic ensemble selection framework using meta-learning},'' \emph{Pattern
  Recognition}, vol.~48, no.~5, pp. 1925--1935, 2015.

\bibitem{souza2019online}
M.~A. Souza, G.~D. Cavalcanti, R.~M. Cruz, and R.~Sabourin, ``Online local pool
  generation for dynamic classifier selection,'' \emph{Pattern Recognition},
  vol.~85, pp. 132--148, 2019.

\end{thebibliography}

\end{document}